\documentclass{article}

\usepackage[backend=bibtex, style=ieee]{biblatex}
\bibliography{bibliography}
\usepackage[utf8]{inputenc} 
\usepackage[T1]{fontenc}    
\usepackage[bookmarks=false]{hyperref}       
\usepackage{url}            
\usepackage{booktabs}       
\usepackage{amsfonts}       
\usepackage{nicefrac}       
\usepackage{microtype}      
\usepackage{amsmath}
\usepackage{amssymb}
\usepackage{subcaption}
\usepackage{multirow}
\usepackage{color,soul}
\usepackage{graphics}
\usepackage[nottoc,notlot,notlof]{tocbibind}
\hypersetup{colorlinks, citecolor=green, filecolor=black, linkcolor=blue, urlcolor=blue}
\usepackage{lipsum}
\usepackage{booktabs}
\usepackage{amsmath}
\usepackage{amssymb}
\usepackage{amsfonts}
\usepackage{pifont}
\usepackage{etoolbox}
\usepackage{xargs}
\usepackage{tabularx}
\newcolumntype{P}[1]{>{\centering\arraybackslash}p{#1}}
\newcolumntype{K}[1]{>{\centering\arraybackslash}p{#1}}
\usepackage[pdftex]{graphicx}
\usepackage{csquotes}
\usepackage[noend]{algpseudocode}
\usepackage{algorithmicx}
\usepackage{algorithm}
\newtheorem{thm}{Theorem}
\newtheorem{hyp}{Hypothesis}
\algdef{SE}[DOWHILE]{Do}{doWhile}{\algorithmicdo}[1]{\algorithmicwhile\ #1}%

\usepackage[preprint]{neurips_2018}

\title{
Bootstrapped Coordinate Search for Multidimensional Scaling}

\author{%
  Efthymios Tzinis \\
  Department of Computer Science \\
  University of Illinois Urbana-Champaign \\
  Champaign, IL 61820 \\
  \texttt{etzinis2@illinois.edu} 
}

\begin{document}

\maketitle

\begin{abstract}

In this work, a unified framework for gradient-free Multidimensional Scaling (MDS) based on Coordinate Search (CS) is proposed. This family of algorithms is an instance of General Pattern Search (GPS) methods which avoid the explicit computation of derivatives but instead evaluate the objective function while searching on coordinate steps of the embedding space. The backbone element of CSMDS framework is the corresponding probability matrix that correspond to how likely is each corresponding coordinate to be evaluated. We propose a Bootstrapped instance of CSMDS (BS CSMDS) which enhances the probability of the direction that decreases the most the objective function while also reducing the corresponding probability of all the other coordinates. BS CSMDS manages to avoid unnecessary function evaluations and result to significant speedup over other CSMDS alternatives while also obtaining the same error rate. Experiments on both synthetic and real data reveal that BS CSMDS performs consistently better than other CSMDS alternatives under various experimental setups.
\end{abstract}

\let\thefootnote\relax\footnotetext{\noindent Code is available at: \url{https://github.com/etzinis/bootstrapped_mds}}


\section{Introduction}
Derivative-free optimization has been an active field of study in the past years. These methods try to optimize an objective function $f:\mathbb{R}^n \rightarrow \mathbb{R}$ by searching over the input space $\mathbb{R}^n$ and evaluate the function on different points. In each step, the algorithm selects the point that produces the steepest decrease on the value of the objective function. On the next iterate, the algorithm will perform the aforementioned process but starting from the point that has been produced by the previous iterate. These methods might be useful in cases where the derivative of the objective function $\nabla f$ is not available, it is extremely noisy or it is hard to compute. As a result, we can directly optimize any objective function $f$ by replacing any gradient-based method by a derivative-free one. Derivative-free optimization methods belong to the class of General Pattern Search (GPS) methods which are known to be able to converge to local minima under mild restrictions of the objective function $f$ \cite{torczon1997convergence}. However, a harsh problem of those methods lies on their inability to scale on multiple dimensions as they scale linearly with the number of dimensions of the search space $n$.  

Multidimensional Scaling (MDS) \cite{ClassicalMDS} has been one of the most renowned dimensionality reduction techniques that tries to find a lower-dimensional manifold $\mathbf{X} \in \mathbb{R}^n$ which are preserving a given similarity matrix $\mathbf{\Delta} \in  \mathbb{R}^{n \times n}$. In essence, classical MDS tries to preserve these pairwise distances $\hat{d_{ij}}$ (or more general dissimilarities) between the $n$ points of the manifold $d_{ij}(\mathbf{X}))$ by minimizing the stress objective function: $\sigma_{raw}^2(\mathbf{X},\hat{\mathbf{D}}) = 
\sum_{i=1}^{n} \sum_{j=1}^{n} (\hat{d_{ij}}-d_{ij}(\mathbf{X}))^2 $, where no closed form solution is available \cite{Kruskal:a}. Conventional methods would try to minimize the aforementioned objective function using gradient based methods. Moreover, the state-of-the-art algorithm for solving the problem is Scaling by Majorizing a Complex Function (SMACOF) \cite{Leeuw77applicationsof} which is minimizing the aforementioned stress by minimizing a function which is greater or equal than the objective. 

Previous approaches have incorporated a simple derivative-free optimization method which is called direct coordinate search or simpler Coordinate Search (CS) search \cite{Hooke:1961:DSS:321062.321069} in order to minimize the stress function which is the objective function for MDS problem \cite{paraskevopoulostzinis2018MDS}. One possible method to reduce the search space of the algorithm would be to naively random sample over the possible dimensions that the algorithm searches over. We are focused on increasing the efficiency of MDS using direct search by using a Bootstrap sampling method \cite{efron1994introduction} on the search space of all the possible directions. In essence, we assimilate the momentum of the optimization movement by keeping a history of the previous selected steepest descent directions and biasing towards selecting them from all the available coordinates. For each iteration we update the dictionary of the probabilities for all the directions by increasing the probability of evaluating a successful coordinate if it provided the minimum alongside all the other coordinates and we reduce the probability of selecting all the others accordingly. One can consider that we try to optimize the function by selecting only a sparse subset of all the available coordinates. 

In this work, we propose a general framework of Coordinate Search (CS) optimization algorithm alternatives for solving MDS. Full-Search Coordinate Search MDS (FS CSMDS) searches over all the possible directions for each iteration and also comes with some theoretical convergence properties. On the contrary, Randomized Coordinate Search MDS (RN CSMDS) will evaluate only on a randomly selected subset of all possible coordinates for evaluating the function which seems to work faster but without any theoretical reassurance of its convergence. Finally, Bootstrapped Coordinate Search MDS (BS CSMDS) does some kind of importance sampling in order to select the subset of the available directions that the objective function would be evaluated and can also be presented in the same general framework like the latter two alternatives. We experiment using synthetic data and we qualitatively inspect how well MDS can create low-dimensional manifolds able to preserve the input dissimilarities for multiple cases of the randomized selection of the optimization coordinate directions. We also provide the same results when trying to reconstruct a nonlinear manifold by using as input the geodesic distances of the input data instead of the euclidean ones. By the latter approach, we effectively resemble the functionality of Isometric Mapping (ISOMAP) \cite{tenenbaum_global_2000}. Furthermore, we try to give a quantitative performance measure using real data from MNIST images \cite{lecun1998mnist} in order to evaluate the efficacy of these low dimensional embeddings under a K-Nearest Neighbor (KNN) classification scheme. This would give us a deeper insight on how well the intrinsic geometry of the data is preserved when we perform dimensionality reduction. All the above would be consistently evaluated for all the cases of: 1) FS CSMDS, 2) RN CSMDS and 3) BS CSMDS. Finally, we compare these three alternatives of CSMDS framework under different configurations in order to assess the efficacy of the proposed Bootstraped Coordinate Search MDS (BS CSMDS).

\section{Preliminaries}
\subsection{Multidimensional Scaling (MDS)}
\label{Background:MDS}

\subsubsection{Classical MDS}
Classical MDS was first introduced in \cite{ClassicalMDS}. MDS can be formalized as: given the matrix $\mathbf{\Delta}$ consisting of pairwise distances or dissimilarities $\{\delta_{ij}\}_{1 \leq i , j \leq N}$ between $N$ points in a high dimensional space, we seek to find a set of points $\{\mathbf{x}_i\}_{i=1}^N$ which lie on the manifold $\mathcal{M} \in \mathbb{R}^{L}$ and their pairwise distances are preserve the given dissimilarities $\{\delta_{ij}\}_{1 \leq i , j \leq N}$ as truthfully as possible. $\mathbf{X}^T \in \mathbb{R}^{L \times N}$ corresponds to the data array contains all points $\mathbf{x}_i \in \mathbb{R}^{L}, \enskip 1 \leq i \leq N$ as its columns.
Ideally, we would like to obtain an embedding dimension $L$ as small as possible in order to reduce as much as possible the dimensionality of the produced manifold $\mathcal{M}$ but not to deviate much from the given dissimilarities $\delta_{ij}$. Usually, euclidean distances are considered $d_{ij}(\mathbf{X})=||\mathbf{x}_i - \mathbf{x}_j||_2 = \sqrt{\sum_{k=1}^L (x_{ik}-x_{jk})^2} $ in the embedded space $\mathbb{R}^{L}$.  

Classical MDS uses a centering matrix $\mathbf{H}=\mathbf{I}_N - \frac{1}{N} \mathbf{1}_N^T \mathbf{1}_N  $ which is effectively a way to subtract the mean of the columns and the rows for each element. Where $\mathbf{1}_N = [1,1,...,1]$ is a vector of ones in $\mathbb{R}^N$ space. By applying the double centering to the Hadamard product of the given dissimilarities, the Gram matrix $\mathbf{B}$ is constructed as follows:
\begin{equation}
\label{eq gram matrix}
\mathbf{B} = - \frac{1}{2}\mathbf{H}^T (\mathbf{\Delta} \odot \mathbf{\Delta}) \mathbf{H} 
\end{equation}
As shown in Ch. 12 \cite{borg_groenen_2005}, Classical MDS minimizes the Strain algebraic criterion, namely:
\begin{equation}
\label{eq Strain}
||\mathbf{X}\mathbf{X}^T-\mathbf{B}||_F^2
\end{equation}
$\mathbf{B}= \mathbf{V} \mathbf{\Lambda}\mathbf{V}^T$ is a symmetric matrix. The embedded points in $\mathbb{R}^L$ are given by the first $L$ positive eigenvalues of $\mathbf{\Lambda}$, namely $\mathbf{X}=\mathbf{V}_L$. As shown in \cite{doi:10.1093/biomet/53.3-4.325} the solution to Classical MDS provides the same result as Principal Component Analysis (PCA) if the latter is applied on the vector in the high dimensional space.
Classical MDS was originally proposed for dissimilarity matrices $\mathbf{\Delta}$ which can be embedded with good approximation accuracy in a low-dimensional Euclidean space as we also consider in this work. However, matrices
which correspond to different spaces such as: Euclidean sub-spaces \cite{10.1007/978-3-642-46900-8_44}, Poincare disks \cite{Poincare:MDS} and constant-curvature Riemannian spaces \cite{lindman1978constant} have also been studied. 

\subsubsection{Metric MDS}
Metric MDS describes contains Classical MDS. Shepard has introduced heuristic methods to enable transformations of the given dissimilarities $\delta_{ij}$ \cite{Shepard1962}, \cite{Shepard1962b} but did not provide any loss function in order to model them \cite{groenen2014past}. Kruskal in \cite{Kruskal:a} and \cite{Kruskal:b} formalized the metric MDS as a least squares optimization problem of minimizing the non-convex Stress-1 function defined below:
\begin{equation}
\label{eq stress1}
\sigma_1(\mathbf{X},\hat{\mathbf{D}}) = 
\sqrt[]{\frac{\sum_{i=1}^{N} \sum_{j=1}^{N} (\hat{d_{ij}}-d_{ij}(\mathbf{X})) }{\sum_{i=1}^{N} \sum_{j=1}^{N} d_{ij}^2( \mathbf{X}) }}
\end{equation}
where matrix $\hat{\mathbf{D}}$ with elements $\hat{d_{ij}}$ represents all the pairs of the transformed dissimilarities $\delta_{ij}$ that are used to fit the embedded distance pairs $d_{ij}(\mathbf{X})$. 

A weighted MDS raw Stress function is defined as: 
\begin{equation}
\label{eq raw_stress}
\sigma_{raw}^2(\mathbf{X},\hat{\mathbf{D}}) = 
\sum_{i=1}^{N} \sum_{j=1}^{N} w_{ij}(\hat{d_{ij}}-d_{ij}(\mathbf{X}))^2 
\end{equation}
\noindent
where the weights $w_{ij}$ are restricted to be non-negative; for missing data the weights are set equal to zero. 
In our work, we consider always $w_{ij}=1, \space \forall 1 \leq i,j \leq N$ where we assume an equal contribution to the Metric-MDS solution for all the elements.  

\subsubsection{SMACOF}
A state-of-the-art algorithm for solving metric MDS is SMACOF stands for Scaling by Majorizing a Complex Function \cite{Leeuw77applicationsof}. By setting $\hat{d_{ij}}=\delta_{ij}$ in raw stress function defined in Equation~\ref{eq raw_stress}, SMACOF minimizes the resulting stress function $\sigma_{raw}^2(\mathbf{X})$.
\begin{equation}
\label{eq smacof_stress}
\sigma^2(\mathbf{X}) = 
\sum_{i=1}^{N} \sum_{j=1}^{N} w_{ij}(\delta_{ij}^2-2 \delta_{ij} d_{ij}(\mathbf{X}) + d_{ij}^2(\mathbf{X})) 
\end{equation}
The algorithm proceeds iteratively and decreases stress monotonically by optimizing a convex function which serves as an upper bound for the non-convex stress function in Equation~\ref{eq smacof_stress}. SMACOF has been extensively described in \cite{borg_groenen_2005} while its convergence for a Euclidean embedded space $\mathbb{R}^L$ has been proven in \cite{deLeeuw1988}.

Let matrices $\mathbf{U}$ and $\mathbf{R(\mathbf{X})}$ be defined element-wise as follows: 
\begin{equation}
\label{eq Moore Penrose}
u_{ij} = 
\left\lbrace
\begin{array}{ll}
      -w_{ij} & i \ne j \\
      \sum_{k \ne i} w_{ik} & i = j
\end{array} 
\right.
\end{equation}
\begin{equation}
\label{eq supporting matrix}
r_{ij} = 
\left\lbrace
\begin{array}{ll}
      -w_{ij}\delta_{ij}d_{ij}^{-1}(\mathbf{X}) & i \ne j, d_{ij}(\mathbf{X}) \ne 0 \\
      0 & i \ne j, d_{ij}(\mathbf{X}) = 0 \\
      \sum_{k \ne i} r_{ik} & i = j
\end{array} 
\right.
\end{equation}
An alternative view of Equation~\ref{eq smacof_stress} is the equivalent quadratic form: 
\begin{equation}
\label{eq smacof_stress_converted}
\sigma^2(\mathbf{X}) = 
\sum_{i=1}^{N} \sum_{j=1}^{N} w_{ij}\delta_{ij}^2
- 2 tr(\mathbf{X}^T \mathbf{R}(\mathbf{X}) \mathbf{X})
+tr(\mathbf{X}^T \mathbf{U} \mathbf{X}) 
\end{equation}
The quadratic is minimized iteratively as follows:
\begin{equation}
\label{eq majorizing func}
T(\mathbf{X}, \hat{\mathbf{X}}^{(k)}) = \sum_{j=1}^{N} w_{ij}\delta_{ij}^2 - 2 tr(\mathbf{X}^T \mathbf{R}(\hat{\mathbf{X}}^{(k)}) \hat{\mathbf{X}}^{(k)}) +tr(\mathbf{X}^T \mathbf{U} \mathbf{X})  
\end{equation}
Now, for $k$-th iteration we obtain the next optimal point:
\begin{equation}
\label{eq SMACOF update}
\hat{\mathbf{X}}^{(k+1)}=\underset{\mathbf{X}}{\operatorname{argmin}}\;{T(\mathbf{X}, \hat{\mathbf{X}}^{(k)})}=\mathbf{U}^{\dagger}\mathbf{R}(\hat{\mathbf{X}}^{(k)}) \hat{\mathbf{X}}^{(k)}
\end{equation}
where $\hat{\mathbf{X}}^{(k)}$ is the estimate of the embedding data matrix $\mathbf{X}$ at the $k$th iteration and $\mathbf{U}^{\dagger}$ is Moore-Penrose pseudo-inverse of $\mathbf{U}$. Noticeably, $\sum_{i=1}^{N} \sum_{j=1}^{N} w_{ij}\delta_{ij}^2$ is constant for all data matrices estimations for each iteration.
At $k$th iteration, the convex majorizing convex function touches the surface of $\sigma$ at the point $\hat{\mathbf{X}}^{(k)}$. By minimizing the quadratic problem in Equation~\ref{eq majorizing func} we find the next update which serves as a starting point for the next iteration $k+1$. The solution to the minimization problem is shown in Equation~\ref{eq SMACOF update}. The algorithm stops when the new update yields a decrease $\sigma^2(\hat{\mathbf{X}}^{(k+1)})-\sigma^2(\hat{\mathbf{X}}^{(k)})$ that is smaller than a threshold value.

\subsection{General Pattern Search (GPS) methods}
\label{Background:GPS}
A wide class of derivative-free algorithms for nonlinear optimization has been studied and analyzed in \cite{Rios2013} and \cite{avriel2003nonlinear}. GPS methods are a subset of these algorithms which do not require the explicit computation of the gradient in each iteration-step. Some GPS algorithms are: the original Hooke and Jeeves direct coordinate search algorithm \cite{Hooke:1961:DSS:321062.321069}, the evolutionary operation by utilizing factorial design \cite{box1957evolutionary} and the multi-directional search algorithm \cite{torczon1989multidirectional}, \cite{doi:10.1137/0801027}. In \cite{torczon1997convergence}, a unified theoretical formulation of GPS algorithms under a common notation model has been presented as well as an extensive analysis of their global convergence properties. 

\subsubsection{GPS formulation}
\label{section: GPS formulation}
GPS methods optimize an objective $f:\mathbb{R}^n \rightarrow \mathbb{R}$:
\begin{equation}
\label{eq General minimization}
\mathbf{x}^*=\underset{\mathbf{x} \in \mathbb{R}^n}{\operatorname{argmin}} \;f(\mathbf{x})
\end{equation}
GPS try to minimize Equation~\ref{eq General minimization}. The general framework for all pattern search methods also contains Coordinate Search methods. A full formalization of the general framework of all GPS methods is given in \cite{torczon1997convergence,dolan2003local}. Firstly, we define the following components:
\begin{itemize}
\item A basis matrix that could be any nonsingular matrix $\mathbf{B} \in \mathbb{R}^{n \times n}$. 
\item A matrix $\mathbf{C}^{(k)}$ for generating all the possible direction for the $k$th iteration of the GPS minimization algorithm 
\begin{equation}
\label{eq GPS generating matrix}
\mathbf{C}^{(k)}=[\mathbf{M}^{(k)} \enskip -\mathbf{M}^{(k)} \enskip \mathbf{L}^{(k)}] = [\Psi^{(k)} \enskip \mathbf{L}^{(k)}]
\end{equation}
\noindent
where the columns of $\mathbf{M}^{(k)} \in \mathbb{Z}^{n \times n}$  form a positive span of $\mathbb{R}^{n}$ and $\mathbf{L}^{(k)}$ contains at least the zero column of the search space $\mathbb{R}^{n}$.  
\item A pattern matrix $\mathbf{P}^{(k)}$ defined as
\begin{equation}
\label{eq GPS pattern matrix}
\mathbf{P}^{(k)}=\mathbf{B}\mathbf{C}^{(k)}=[\mathbf{B}\mathbf{M}^{(k)} \enskip -\mathbf{B}\mathbf{M}^{(k)} \enskip \mathbf{B}\mathbf{L}^{(k)}]
\end{equation}
where the submatrix $\mathbf{B}\mathbf{M}^{(k)}$ forms a basis of $\mathbb{R}^{n}$. 
\end{itemize}

In each iteration $k$, the set of steps $\{\mathbf{s}_i^{(k)}\}_{i=1}^{m}$ are generated by the pattern matrix $\mathbf{P}^{(k)}$: 
\begin{equation}
\label{eq GPS trial step}
\mathbf{s}_i^{(k)}=\Delta^{(k)} \mathbf{p}_{i}^{(k)}, \enskip \mathbf{P}^{(k)} = [\mathbf{p}_{1}^{(k)},...,\mathbf{p}_{m}^{(k)}] \in \mathbb{R}^{n \times m}
\end{equation}
\noindent
where $\mathbf{p}_{i}^{(k)}$ is the $i$th column of defines the direction of the new step, while $\Delta^{(k)}$ configures the length towards this direction. If pattern matrix $\mathbf{P}^{(k)}$ contains $m$ columns, then $m \geq n + 1$ in order to positively span the search space $\mathbb{R}^{n}$. A new trial point of GPS algorithm would be $\mathbf{x}_i^{(k+1)}=\mathbf{x}^{(k)}+\mathbf{s}_i^{(k)}$ where we evaluate the value of the function $f$ which we seek to minimize. A successful step would mean a further minimization of the objective function would mean $f$, i.e., $f(\mathbf{x}^{(k)}+\mathbf{s}_i^{(k)}) > f(\mathbf{x}^{(k+1)})$. A pseudo-code for all GPS methods is presented in Algorithm \ref{alg: GPS algorithm}.
\begin{algorithm}
\caption{General Pattern Search (GPS)}
\label{alg: GPS algorithm}
\begin{algorithmic}[1] 
\Procedure{GPS\_SOLVER}{$\mathbf{x}^{(0)}$, $\Delta^{(0)}$, $\mathbf{C}^{(0)}$, $\mathbf{B}$}
\State $k=-1$
\Do
	\State $k = k+1$
    \State $\mathbf{s}^{(k)} =$ EXPLORE\_MOVES($\mathbf{B}\mathbf{C}^{(k)}$, $\mathbf{x}^{(k)}$, $\Delta^{(k)}$) \label{func: GPS exploratory moves}
    \State $\rho^{(k)}=f(\mathbf{x}^{(k)}+\mathbf{s}^{(k)})-f(\mathbf{x}^{(k)})$
    \If{$\rho^{(k)} < 0$}
    	\State $\mathbf{x}^{(k+1)}=\mathbf{x}^{(k)} + \mathbf{s}^{(k)}$
        \Comment{Successful iteration}
    \Else 
    	\State $\mathbf{x}^{(k+1)}=\mathbf{x}^{(k)}$
        \Comment{Unsuccessful iteration}
    \EndIf
    \State $\Delta^{(k+1)}, \mathbf{C}^{(k+1)}=$ UPDATE($\mathbf{C}^{(k)}$, $\Delta^{(k)}$,  $\rho^{(k)}$) \label{func: GPS update ck and dk}
    
\doWhile{convergence criterion is not met}
\EndProcedure
\end{algorithmic}
\end{algorithm}

Initially, we select $\mathbf{x}^{(0)} \in \mathbb{R}^n$ and a positive step length parameter $\Delta^{(0)} > 0$. In each iteration $k$, we explore a set of possible steps defined by the $\texttt{EXPLORE\_MOVES}()$ subroutine at line~\ref{func: GPS exploratory moves} of the algorithm. Pattern search methods mainly differ on the heuristics used for the selection of exploratory moves that they evaluate the function $f$ on. If a new exploratory point lowers the value of the 
function $f$, iteration $k$ is considered successful and the starting point of the next iteration is updated $\mathbf{x}^{(k+1)}=\mathbf{x}^{(k)} + \mathbf{s}^{(k)}$. Otherwise, if at a certain iteration we cannot obtain any successful step then the algorithm can produce the zero-step point. The step length parameter $\Delta^{(k)}$ is modified by the $\texttt{UPDATE}()$ subroutine. For successful iterations, i.e.,  $\rho^{(k)} < 0$, the step length is forced to increase in a way that can be described as follows:
\begin{equation}
\label{eq Delta Increase}
\begin{split}
& \Delta^{(k+1)} = \lambda^{(k)} \Delta^{(k)}, \enskip \lambda^{(k)}  \in \Lambda = \{\tau^{w_1},...,\tau^{w_{|\Lambda|}}\} \\
& \tau > 1, \enskip \{w_1,...,w_{|\Lambda|}\} \subset \mathbb{N}, \enskip |\Lambda| < +\infty 
\end{split}
\end{equation}
where $\tau$ and $w_i$ are predefined constants that are used for the $i$th successive successful iteration.
For unsuccessful iterations the step length parameter is decreased, i.e., $\Delta^{(k+1)} \leq \Delta^{(k)}$ as follows: 
\begin{equation}
\label{eq Delta Decrease}
\Delta^{(k+1)} = \theta \Delta^{(k)}, \enskip \theta = \tau^{w_0}, \enskip \tau > 1, \enskip w_0 < 0, 
\end{equation}
where $\tau$ and the negative integer $w_0$ determine the fixed ratio of step reduction. Generating matrix $\mathbf{C}^{(k+1)}$ could be also updated for unsuccessful/successful iterations in order to contain more/less search directions, respectively.

\subsubsection{GPS Convergence}
\label{section: GPS Convergence}
Important convergence properties have been shown in \cite{torczon1997convergence, dolan2003local, doi:10.1137/S1052623496300507, doi:10.1137/0728030, doi:10.1137/S1052623497331373} for any GPS method that can be described by the previously defined framework.
\begin{hyp}[Weak Hypothesis on Exploratory Moves]
\label{hyp: mild exploratory moves} 
The subroutine $\texttt{EXPLORE\_MOVES}()$ as defined in Algorithm~\ref{alg: GPS algorithm}, line~\ref{func: GPS exploratory moves} guarantees the following:
\begin{itemize}
\item The exploratory step direction for iteration $k$ is selected from the columns of the pattern matrix $\mathbf{P}^{(k)}$ as defined in Equation~\ref{eq GPS trial step} and the exploratory step length is ${\Delta^{(k)}}$ as defined in Equations \ref{eq Delta Increase}, \ref{eq Delta Decrease}.
\item If among the exploratory moves $\mathbf{a}^{(k)}$ at iteration $k$ selected from the columns of the matrix $\Delta^{(k)}\mathbf{B}[\mathbf{M}^{(k)}  -\mathbf{M}^{(k)}]$ exist at least one move that leads to a successful iteration, i.e., $f(\mathbf{x}^{(k)}+\mathbf{a}) < f(\mathbf{x}^{(k)})$, then the $\texttt{EXPLORE\_MOVES}()$  subroutine will return a move $\mathbf{s}^{(k)}$ such that 
$f(\mathbf{x}^{(k)}+\mathbf{s}^{(k)})<f(\mathbf{x}^{(k)})$.
\end{itemize}
\end{hyp} 
Hypothesis \ref{hyp: mild exploratory moves} enforces some mild constraints on how the exploratory moves would be produced by Algorithm \ref{alg: GPS algorithm}, line~\ref{func: GPS exploratory moves}. Essentially, the suggested step $\mathbf{s}^{(k)}$ must be derived from the pattern matrix $\mathbf{P}^{(k)}$. Moreover, the algorithm needs to provide a simple decrease for the objective function $f$ at every step. Specifically, the only way to accept an unsuccessful iteration would be if none of the steps from the columns of the matrix $\Delta^{(k)}\mathbf{B}[\mathbf{M}^{(k)}  -\mathbf{M}^{(k)}]$ lead to a decrease of the objective function $f$. 

Based on the aforementioned Hypothesis, a GPS method can enjoy some theoretical convergence guarantees which are stated rigorously in Theorem. \ref{theorem: Mild Convergence} as follows:
\begin{thm}
\label{theorem: Mild Convergence}
Let $L(\mathbf{x}^{(0)})=\{\mathbf{x}: f(\mathbf{x}) \leq f(\mathbf{x}^{(0)})\}$ be closed and bounded and $f$  continuously differentiable on a neighborhood of $L(\mathbf{x}^{(0)})$, namely on the union of the open balls $\underset{{\mathbf{a} \in L(\mathbf{x}^{(0)})}}{\bigcup} B(\mathbf{a}, \eta)$ where $\eta > 0$. If a GPS method is formulated as described in Section \ref{section: GPS formulation} and Hypothesis \ref{hyp: mild exploratory moves} holds then for the sequence of iterations $\{\mathbf{x}^{(k)}\}$ produced by Algorithm~\ref{alg: GPS algorithm} 
\[\underset{k \rightarrow +\infty}{\lim} \inf \; ||\nabla f(\mathbf{x}^{(k)})|| = 0 \]
\end{thm}
For the proof of this Theorem we refer the reader to \cite{torczon1997convergence}.

However, as shown in \cite{Audet2004} one can construct a continuously differentiable objective function and a GPS method with infinite many limit points with non-zero gradients and thus even Theorem. \ref{theorem: Mild Convergence} holds, the convergence of $||\nabla f(\mathbf{x}^{(k)})||$ is not assured. 

\section{Coordinate Search Multidimensional Scaling (CSMDS)}
Coordinate search is a specific case of GPS methods that have been described before and thus it can be formulated under the aforementioned GPS framework (see Section \ref{section: GPS formulation}). In this section we try to unify the three alternatives of CSMDS: 1) Full-Search Coordinate Search MDS (FS CSMDS), 2) Randomized Coordinate Search MDS (RN CSMDS) and 3) Bootstrapped Coordinate Search MDS (BS CSMDS) under a common family of algorithms which encapsulate a probability of selecting one dimension in order to evaluate the objective function. this family of algorithms can be directly restated as specific examples of GPS methods. Moreover, we discuss about the complexity of these algorithms in each case and we restate the convergence reassurance of FS CSMDS as firstly introduced in \cite{paraskevopoulostzinis2018MDS}.

\subsection{Unified CSMDS Algorithm}
\label{MDS:AlgorithmDescription}

In accordance with \cite{paraskevopoulostzinis2018MDS}, we consider a derivative-free optimization method for MDS. The input to CSMDS is a $N \times N$ target dissimilarity matrix $\mathbf{T}$ and the target dimension $L$ of the embedding space. The algorithm of the unified CSMDS algorithms is shown in Algorithm~\ref{a:proposed}.

First of all, we initialize randomly $\mathbf{X}^{(0)} = [ \mathbf{x}_1^{(0)}, \mathbf{x}_2^{(0)}, ..., \mathbf{x}_N^{(0)} ] \in \mathbb{R}^{N \times L}$ and we compute the euclidean distance matrix $\mathbf{D}^{(0)}$ of the embedded space. Thus, $d_{ij}^{(0)}$ represents the Euclidean distance between vectors $\mathbf{x}_i^{(0)}$ and $\mathbf{x}_j^{(0)}$ of $\mathbf{X}^{(0)}$. The approximation error $e^{(0)} = f(\mathbf{T}, \mathbf{D}^{(0)})$ is the element-wise mean squared error (MSE) between the two matrices. The objective function which is optimized for all CSMDS variants is shown below and is the stress function for MDS:
\begin{equation}
 f(\mathbf{T}, \mathbf{D}) = ||\mathbf{T} - \mathbf{D}||_{F}^2 = \sum_{i = 1}^{N}\sum_{j=1}^{N}(t_{ij} - d_{ij})^2, \enskip \; \mbox{where} \; \mathbf{T}, \mathbf{D}\in \mathbb{R}^{N \times N}
\end{equation}
We introduce the notation of matrix $\tilde{\mathbf{P}}^{(k)} \in [0,1]^{N\times2L}$ that contains the probabilities of all the possible coordinates to be searched for $k$th iteration. Specifically, $\tilde{p}^{(k)}_{ij} \in [0,1]$, ($1\leq i \leq N$ and $1\leq j \leq 2L$) is the probability of evaluating the objective function for point $i$ towards coordinate with index $j$ at $k$th iteration. Because we consider $L$ dimensions of the embedding space we search over all positive and negative coordinates of this space on a certain radius $r^{(k)}$. For example in the case of Full-Search CSMDS we will have $\tilde{p}^{(k)}_{ij} = 1 \enskip \forall i,j$ because we always evaluate the function over all the possible directions for every iteration. Consequently, this probability matrix will not be updated over different epochs in both cases of Full-Search (FS) and Randomized (RN) CSMDS. We also introduce two more variables $p_{th}$ and $p_{a}$ which would be the threshold of the probability for all the values of matrix $\tilde{\mathbf{P}}^{(k)}$ and the step of the update towards the probabilities for all the coordinates when the matrix is updated, respectively. So in any case we will have  $\tilde{p}^{(k)}_{ij} \geq p_{th}$. In the case of the proposed BS CSMDS, when a successful coordinate step $s^{*}_x$ is evaluated for point $x$ (the coordinate that produces the maximum minimization of the objective function) we would enhance the probability of this coordinate for the next iterates by $\tilde{p}^{(k+1)}_{x s^{*}_x } = \tilde{p}^{(k)}_{ x s^{*}_x} + p_{a}$ and we will reduce the probabilities of selecting any of the other non-optimal coordinates by the same amount.
\begin{algorithm}
\caption{Unified CSMDS}
\label{a:proposed}
\begin{algorithmic}[1] 
\Procedure{CSMDS}{$\mathbf{T}$, $L$, $\tilde{\mathbf{P}}^{(0)}$, $r^{(0)}$, $\delta$, $\epsilon$, $p_{th}$, $p_a$}
    \State $k$ $\gets$ $0$
    \Comment{k is the number of epochs}
	\State $\mathbf{X}^{(k)}$ $\gets$ UNIFORM($N \times L$)
    \State $\mathbf{D}^{(k)}$ $\gets$ DISTANCE\_MATRIX($\mathbf{X}^{(k)}$)
    \State $e^{(k)}$ $\gets$ $||\mathbf{T} - \mathbf{D}||_{F}^2$
    \State $e^{(k-1)}$ $\gets$ $+\infty$
    \State $r^{(k)}$ $\gets$ $r^{(0)}$
    \While{$r^{(k)} > \delta$}
        \If{$e^{(k-1)} - e^{(k)} \leq \epsilon \cdot e^{(k)}$} \label{al:eps}
        	\State $r^{(k)}$ $\gets$ $\frac{r^{(k)}}{2}$
        \EndIf
    	
		\ForAll{$x \in \mathbf{X}^{(k)}$}
		    \State $\mathcal{S}_x$ $\gets$ SEARCH\_COORDINATES($r^{(k)}$, $x$, $L$,   $\tilde{\mathbf{P}}^{(k)}$)
			\State $\mathbf{X^{*}}, s^{*}_x, e^{*}$ $\gets$ OPTIMAL\_MOVE($\mathbf{X}^{(k)}$, $x$, $\mathcal{S}_x$, $e^{(k)}$) \label{al:move_line}
			\State $e^{(k-1)} \gets e^{(k)}$
			\State $e^{(k)}$ $\gets$ $e^*$
			\State $\tilde{\mathbf{P}}^{(k+1)}$ $\gets$ UPDATE\_PROBABILITIES($e^{(k-1)}$, $e^{(k)}$, $x$, $s^{*}_x$,  $\tilde{\mathbf{P}}^{(k)}$, $L$, $p_{th}$, $p_a$)
        	\State $\mathbf{X}^{(k)}$ $\gets$ $\mathbf{X^*}$
        \EndFor
        \State $k = k + 1$
	\EndWhile
\EndProcedure
\end{algorithmic}
\end{algorithm}
In other words, Algorithm \ref{a:proposed} minimizes the objective function by iterating over all the points for each epoch and searches over some coordinates. For the given set of all the coordinates, the optimal direction (the coordinate that yields the steepest descent for the objective function) is taken and then this point on the embedding space is updated. The resulting error $e^*$ is computed after performing the optimal move for each point of the data matrix $\mathbf{X}^{(k)}$. If the error does not relatively decrease more than a specified constant $\epsilon > 0$, namely, $e^{(k)} - e^* < \epsilon \cdot e^{(k)}$, we halve the search radius and proceed to the next epoch. The process stops when the search radius $r$ becomes smaller than another positive constant $\delta > 0$. Depending on the variant of CSMDS framework we might need to update the probability matrix $\tilde{\mathbf{P}}^{(k)}$ by favoring the probability of a successful step $s^{*}_x$ over all the other coordinates. In the next sections we describe all these procedures extensively.   

\subsection{Searching Coordinates}

Following the initialization steps, in each epoch (iteration), we consider the surface of a hypersphere of radius $r^{(k)}$ around each point $\textbf{x}$. The possible search directions lie on the surface of a hypersphere along the orthogonal basis of the space, e.g., in the case of $3$-dimensional space along the directions $\pm x,\pm y,\pm z$ on the sphere. Moreover, the no movement would also be considered a possible direction in the case that no other directions are actually producing a decrease for the objective function. This would create the set of all possible search coordinates $\mathcal{S}_{all}= \left\{ \textbf{s}^- \: \textbf{s}^- \in r \cdot \mathbf{I}_{L }  \right\} \cup \left\{ \textbf{s}^+ : \textbf{s}^+ \in -r \cdot \mathbf{I}_{L } \right\} $. Where $\mathbf{s} \in \mathbf{A}$ denotes that vector $\textbf{s}$ is a column of matrix $\mathbf{A}$. Now we create the actual set of all the coordinates that would be evaluated for point $\textbf{x}$, namely $\mathcal{S}_x$ by appending any direction $\textbf{s}$ on the final set with probability $\tilde{p}^{(k)}_{x s }$. This process is summarized in Algorithm~\ref{a:search_directions}:
\begin{algorithm}
\caption{Retrieve Search Coordinates to be Evaluated}
\label{a:search_directions}
\begin{algorithmic}[1] 
\Function{SEARCH\_COORDINATES}{$r^{(k)}$, $x$, $L$,   $\tilde{\mathbf{P}}^{(k)}$}
	\State $\mathcal{S}_x$ $\gets$ $\left\{ \textbf{s}^- : \textbf{s}^- \in r \cdot \mathbf{I}_{L },  \enskip w.p. \enskip \tilde{p}^{(k)}_{x s^- } \right\} \cup \left\{ \textbf{s}^+ : \textbf{s}^+ \in -r \cdot \mathbf{I}_{L }, \enskip w.p. \enskip \tilde{p}^{(k)}_{x s^+ } \right\} $
    \State \Return $\mathcal{S}_x$
\EndFunction
\end{algorithmic}
\end{algorithm}

\subsection{Move Alongside the Coordinate that Produces the Maximum Decrease}

Each point is moved greedily along the dimension that produces the minimum error or no error at all considering the zero step direction $\mathbf{0}$.
Algorithm \ref{a:move} finds the optimal step from the set of coordinates $\mathcal{S}_x$ that minimizes the most the error function $e^{(k)} = f(\mathbf{T},\mathbf{D}^{(k)})$ for each point $\tilde{x}$. The data matrix $\mathbf{X}$ is updated by updating only the corresponding point (which is a row vector of the data matrix). We denote by $\mathbf{X}^{(k)}_{a}$ the row of the data matrix $\mathbf{X}^{(k)}$ which corresponds to the point $\textbf{a}$. Similarly, when we evaluate a specific coordinate movement $\textbf{s}$ out of the set of possible directions $\mathcal{S}_x$ then $s$ would denote the subscript of this coordinate.
\begin{algorithm}
\caption{Find optimal move for a point}
\label{a:move}
\begin{algorithmic}[1] 
\Function{OPTIMAL\_MOVE}{$\mathbf{X}^{(k)}$, $x$, $\mathcal{S}_x$, $e^{(k)}$}
    \State $\mathbf{\tilde{X}}$ $\gets$ $\mathbf{X}^{(k)}$
    \State $\mathbf{X}^*$ $\gets$ $\mathbf{X}^{(k)}$
    \State $s^{*}_x$ $\gets$ $0$
	\State $e^* \gets e^{(k)}$ \Comment{This corresponds to zero-step movement}
	\ForAll{$\textbf{s} \in \mathcal{S}_x$}
    	\State $\mathbf{\tilde{X}}_x$ $\gets$ $\mathbf{X}^{(k)}_x + \textbf{s}$
    	\Comment{Update $x$ row of $\mathbf{\tilde{X}}$ with the new step}
    	\State $\mathbf{\tilde{D}}$ $\gets$ DISTANCE\_MATRIX($\mathbf{\tilde{X}}$) \label{al:distance_matrix}
    	\State $\tilde{e}$ $\gets$ $||\mathbf{T} - \mathbf{\tilde{D}}||_{F}^2$
    	\If{$\tilde{e} < e^*$} \Comment{If step $\textbf{s}$ produces lower error then update}
    	    \State $\mathbf{X}^*$ $\gets$ $\mathbf{\tilde{X}}$
    	    \State $s^{*}_x$ $\gets$ $s$
    		\State $e^*$ $\gets$ $\tilde{e}$
  		\EndIf
	\EndFor
    \State \Return $\textbf{X}^*, s^{*}_x, e^*$
\EndFunction
\end{algorithmic}
\end{algorithm}

\subsection{Updating the Probability Matrix}
\label{MDS:UpdateProbMatrix}

The step for each point is being selected greedily along the dimension that produces the minimum error or no error at all considering the zero step direction $\mathbf{0}$.
Algorithm \ref{a:move} finds the optimal step from the set of coordinates $\mathcal{S}_x$ that minimizes the most the error function $e^{(k)} = f(\mathbf{T},\mathbf{D}^{(k)})$ for each point $\tilde{x}$. The data matrix $\mathbf{X}$ is updated by updating only the corresponding point (which is a row vector of the data matrix). We denote by $\mathbf{X}^{(k)}_{a}$ the row of the data matrix $\mathbf{X}^{(k)}$ which corresponds to the point $\textbf{a}$. Similarly, when we evaluate a specific coordinate movement $\textbf{s}$ out of the set of possible directions $\mathcal{S}_x$ then $s$ would denote the subscript of this coordinate.
\begin{algorithm}
\caption{Updating Probability Matrix for a Specified Point}
\label{a:probabilityupdate}
\begin{algorithmic}[1] 
\Function{UPDATE\_PROBABILITIES}{$e^{(k-1)}$, $e^{(k)}$, $x$, $s^{*}_x$,  $\tilde{\mathbf{P}}^{(k)}$, $L$, $p_{th}$, $p_a$}
    \State $\tilde{\mathbf{P}}$ $\gets$ $\tilde{\mathbf{P}}^{(k)}$
    \If{$e^{(k)} < e^{(k-1)}$} \Comment{If we have a successful step $s^{*}_x$ for point $x$}
        \State $\tilde{\mathbf{P}}_{x s^{*}_x}$ $\gets$  = $\operatorname{min} \left\{ \tilde{\mathbf{P}}^{(k)}_{ x s^{*}_x} + 2p_{a}, 1 \right\}$ \Comment{Increase probability of the best coordinate}
        \State $\mathcal{S}_{all}$ $\gets$ $= \left\{ \textbf{s}^- \: \textbf{s}^- \in r \cdot \mathbf{I}_{L }  \right\} \cup \left\{ \textbf{s}^+ : \textbf{s}^+ \in -r \cdot \mathbf{I}_{L } \right\}$
        \ForAll{$\textbf{s} \in \mathcal{S}_{all} $}
    	    \State $\tilde{\mathbf{P}}_{x s}$ $\gets$  = $\operatorname{max} \left\{ \tilde{\mathbf{P}}^{(k)}_{ x s} - p_{a}, p_{th} \right\}$ \Comment{Reduce the probability of all coordinates}
		\EndFor
  	\EndIf
    \State \Return $\tilde{\mathbf{P}}$
\EndFunction
\end{algorithmic}
\end{algorithm}
\subsection{GPS Formulation of CSMDS}
\label{MDS:GPSFormulation}

CSMDS can be expressed by using the unified GPS formulation introduced in Section \ref{Background:GPS}. In accordance with the authors in \cite{paraskevopoulostzinis2018MDS}, who managed to express the specific case of Full-Search CSMDS as an instance of GPS methods, we do the same for the general framework of CSMDS. 

To begin with, the problem of MDS is restated in a vectorized form which spans all the possible coordinates considered for all points. We use matrix $\mathbf{\Delta}$ with elements $\{\delta_{ij}\}_{1 \leq i, j \leq N}$ as the dissimilarities between $N$ points in the high dimensional space. The set of points $\{\mathbf{x}_i\}_{i=1}^N$ lie on the low dimensional manifold $\mathcal{M} \in \mathbb{R}^L$ and form the column set of matrix $\mathbf{X}^T$. The embedded data matrix $\mathbf{X} \in \mathbb{R}^{N \times L}$ will be now vectorized as an one column vector as shown next: 
\begin{equation}
\label{eq Vectorized Variable}
\begin{gathered}
\mathbf{x}_i = [x_{i1},...,x_{iL}]^T \in \mathbb{R}^{L} , 1 \leq i \leq N \\
\mathbf{z} = vec(\mathbf{X}^T) = [x_{11},...,x_{1L},...,x_{N1},...,x_{NL}]^T
\end{gathered}
\end{equation}
Now our new variable $\mathbf{z}$ lies in the search space $\mathbb{R}^{N \cdot L}$. The distance between any two points $\mathbf{x}_i$ and $\mathbf{x}_j$ of the manifold $\mathcal{M}$ remains the same but is now expressed as a function of the vectorized variable $\mathbf{z}$. Namely, $d_{ij}(\mathbf{X})=||\mathbf{x}_i - \mathbf{x}_j||= \sqrt{\sum_{k=1}^L (x_{ik}-x_{jk})^2}=d_{ij}(\mathbf{z})$. The equivalent objective function to minimize $g$ is the MSE between the given dissimilarities $\delta_{ij}$ and the euclidean distances $d_{ij}$ in the low dimensional manifold $\mathcal{M}$ as defined next:
\begin{equation}
\label{eq MSE of the vectorized var}
g(\mathbf{z}) = ||\mathbf{D}_{\mathbf{z}} - \mathbf{\Delta}||_{F}^2 = \sum_{i = 1}^{N}\sum_{j=1}^{N}(d_{ij}(\mathbf{z}) - \delta_{ij})^2, \enskip \mathbf{z} \in \mathbb{R}^{N \cdot L}
\end{equation}
Consequently, the initial MDS is now expressed as an unconstrained non-convex optimization problem which that tries to mimize the objective function $g$ over the search space of $\mathbb{R}^{N \cdot L}$ (Equation \ref{eq Our algorithm's minimization}). Thus, the degrees of freedom for our solution are $L\cdot N$ corresponding to the $L$ coordinates for all $N$ points on the manifold $\mathcal{M}$.
\begin{equation}
\label{eq Our algorithm's minimization}
\mathbf{z}^*=\underset{\mathbf{z} \in \mathbb{R}^{N \cdot L}}{\min}g(\mathbf{z})
\end{equation}
CSMDS is now expressed in an equivalent way by utilizing the auxiliary variable $\mathbf{z}$. Next, we match each epoch of our initial algorithm with an iteration of a GPS method. Therefore, the steps which are produced by CSMDS would form a sequence of points $\{\mathbf{z}^{(k)}\}$ in the new search space $\mathbb{R}^{N \cdot L}$. Moreover, we define the matrices $\mathbf{B}, \mathbf{C}^{(k)}, \mathbf{P}^{(k)}$ for the general framework of CSMDS in an equivalent way to Equations~\ref{eq GPS generating matrix}, \ref{eq GPS pattern matrix}. The choice of our basis matrix $\mathbf{B}$ is the corresponding identity matrix of our search space as shown in Equation~\ref{eq Our algorithm's B}. 
\begin{equation}
\label{eq one hot vector}
\mathbf{e}_i = \underset{index \enskip i}{[0,..,\underbrace{1},...,0]}^T, 1 \leq i \leq N \cdot L 
\end{equation}
\begin{equation}
\label{eq Our algorithm's B}
\mathbf{B} = \mathbf{I}_{N \cdot L} = [\mathbf{e}_1,...,\mathbf{e}_{N \cdot L}]
\end{equation}
Because the columns of $\mathbf{I}_{N \cdot L}$ identity matrix span positively the search space $\mathbb{R}^{N \cdot L}$, we consider as $\mathbf{M}^{(k)}_{\tilde{\mathbf{P}}^{(k)}}$ the matrix which is constructed by all the available positive coordinate directions but with respect to the probability matrix $\tilde{\mathbf{P}}^{(k)}$. Because there is one to one mapping from the auxiliary variable $\mathbf{z}$ to any point of the initial data matrix $\mathbf{X}^T$ we can construct the set of all the considered directions using the subscript of each point $\textbf{x}_i$ and all available positive $L$ coordinates of the embedding space. Formally, we can construct the set of all the coordinate directions that would be considered on $k$th epoch as: 
\begin{equation}
\begin{gathered}
\mathcal{S}_+^{(k)} = \bigcup_{i=1}^{N} \left\{ \textbf{e}_{i \cdot L + s} : \textbf{e}_{i \cdot L + s}  \in  \mathbf{I}_{N \cdot L},  \enskip w.p. \enskip \tilde{\mathbf{P}}^{(k)}_{i s } \right\}  \\
s^- = s +L \\
\mathcal{S}_-^{(k)} = \bigcup_{i=1}^{N} \left\{- \textbf{e}_{i \cdot L + s^-} : - \textbf{e}_{i \cdot L + s^-}  \in - \mathbf{I}_{N \cdot L},  \enskip w.p. \enskip \tilde{\mathbf{P}}^{(k)}_{i s^- } \right\}
\end{gathered}
\end{equation}
where $s$ denotes the subscript of a specific positive coordinate of the space $\mathbb{R}^L$ and $i$ is the subscript that corresponds to the point $\textbf{x}_i$ or equivalently the $i$th row of the data matrix $\textbf{X}$. The subscript $i\cdot L + s$ would correspond to the $(i,s)$ entry of the probability matrix $\tilde{\mathbf{P}} \in [0,1]^{N \times 2L}$. Similarly, for the case of negative coordinates we would have that the corresponding probabilities would lie on the second half of the columns of the probability matrix $\tilde{\mathbf{P}}^{(k)}$ and this is why we need the notation $s^- = s +L$ for the $s$th negative coordinate direction. 
Now the matrices $\mathbf{M}^{(k)}_{\tilde{\mathbf{P}}^{(k)}}$ and $-\mathbf{M}^{(k)}_{\tilde{\mathbf{P}}^{(k)}}$ are constructed by concatenating all the vectors of the sets $\mathcal{S}_+^{(k)}$ and $\mathcal{S}_-^{(k)}$, respectively. This means that we get all the possible coordinate steps alongside the unit coordinate vectors of $\mathbb{R}^{N \cdot L}$. In Equation \ref{eq Our algorithm's Psik} matrix $\Psi^{(k)}$ concatenates the aforementioned matrices in order to incorporate both negative and positive coordinate steps. Nevertheless, our generating matrix $\hat{\mathbf{C}}^{(k)}$ also comprises the zero step vector $\textbf{0} \in \mathbb{R}^{N \cdot L}$ which would be a constant vector-matrix $\mathbf{L}^{(k)} = \hat{\mathbf{L}} = \textbf{0}$.
\begin{equation}
\label{eq Our algorithm's Psik}
\begin{gathered}
\Psi^{(k)} = [\mathbf{M}^{(k)}_{\tilde{\mathbf{P}}^{(k)}} \enskip -\mathbf{M}^{(k)}_{\tilde{\mathbf{P}}^{(k)}}]
\end{gathered}
\end{equation}
\begin{equation}
\label{eq Our algorithm's Lambdak}
\begin{gathered}
\mathbf{L}^{(k)} \equiv \hat{\mathbf{L}} = \textbf{0}
\end{gathered}
\end{equation}
According to Equations~\ref{eq Our algorithm's Psik}, \ref{eq Our algorithm's Lambdak}, we construct the full pattern matrix $\mathbf{P}^{(k)}$ in Equation~\ref{eq Our algorithm's Pk} in a similar way to Equation~\ref{eq GPS pattern matrix}. For our algorithm the pattern matrix is identical to our generating matrix $\mathbf{C}^{(k)}$. Conceptually, the generating matrix $\mathbf{C}^{(k)}$ contains all the possible exploratory moves which would be considered in order to evaluate the objective function $g$ on but always with respect to the probability matrix $\tilde{\mathbf{P}}^{(k)}$ which controls how probable is each coordinate to be evaluated for a certain epoch.
\begin{equation}
\label{eq Our algorithm's Pk}
\begin{gathered}
\mathbf{C}^{(k)}=[\Psi^{(k)} \enskip \hat{\mathbf{L}}]=[\mathbf{M}^{(k)}_{\tilde{\mathbf{P}}^{(k)}} \enskip -\mathbf{M}^{(k)}_{\tilde{\mathbf{P}}^{(k)}} \enskip \hat{\mathbf{L}}] \\
\mathbf{P}^{(k)}  \equiv \mathbf{C}^{(k)}
\end{gathered}
\end{equation}
Recalling the notation of Section \ref{section: GPS formulation}, $\hat{\mathbf{s}}^{(k)}$ is the step which is returned from our exploratory moves subroutine at $k$th iteration. For successful iterates that produce a decrease on the objective function $g(\mathbf{z}^{(k)}+\hat{\mathbf{s}}^{(k)})<g(\mathbf{z}^{(k)})$ we do not further increase the length of our moves by limiting $\Lambda = \{1\}$ as follows:  
\begin{equation}
\label{eq Our algorithm's Delta Increase}
\Delta^{(k+1)} = \Delta^{(k)}, \enskip \; \mbox{if} \; \enskip f(\mathbf{z}^{(k)}+\hat{\mathbf{s}}^{(k)})<f(\mathbf{z}^{(k)}) 
\end{equation}
On the other side, for unsuccessful iterations $g(\mathbf{z}^{(k)}+\hat{\mathbf{s}}^{(k)}) \geq g(\mathbf{z}^{(k)})$ we halve the distance by a factor of $2$ by setting $\theta = \frac{1}{2}$ as it is shown next:
\begin{equation}
\label{eq Our algorithm's Delta Decrease}
\Delta^{(k+1)} =  \frac{1}{2} \Delta^{(k)}, \enskip \; \mbox{if} \; \enskip f(\mathbf{z}^{(k)}+\hat{\mathbf{s}}^{(k)})>=f(\mathbf{z}^{(k)}) 
\end{equation}
Another way to understand CSMDS under GPS methods formulation is the following: In each iteration, we acquire all the possible coordinate directions with respect to the probability matrix $\tilde{\mathbf{P}}^{(k)}$. Each entry of this matrix specifies the probability of a Bernoulli random variable to be successful and this is equal to the probability that a specific coordinate of the search space $\mathbb{R}^{N \cdot L}$ to be considered for evaluation. After that, the algorithm proceeds by moving towards this direction and updates the probability matrix by increasing the probability of this coordinate for the next iterates and reduces the probabilities of all the other coordinates. Otherwise, if no direction produces there is no update in the embedding data matrix $\textbf{X}$ and the radius of the search is reduced in half. This approach provides a potential one-hot vector as described in Equation~\ref{eq one hot vector} if the iterate is successful or otherwise, the zero vector $\mathbf{0} \in \mathbb{R}^{N \cdot L}$ as the step update. The final direction vector $\hat{\mathbf{s}}^{(k)}$ for $k$th iteration is computed by summing these one-hot or zero vectors for all the point updates. At the $k$th iteration, the movement would be given by a scalar multiplication of the step length parameter $\Delta^{(k)}$ with the final direction vector in a similar way as defined in Equation~\ref{eq GPS trial step}. A one-hot vector that corresponds to a specific point would correspond to a simple decrease for the objective function $g$ or in the worst case it would represent a zero movement in the search space $\mathbb{R}_{N \cdot L}$. The movement alongside the optimal coordinate from all points is an associative operation and thus the summation of all these vectors would result in an equivalent update for the auxiliary variable $\textbf{z}$. In other words, accumulating all best coordinate steps for each point $\{\mathbf{x}_i\}_{i=1}^N$ and performing the movement at the end of the $k$th iteration (as GPS method formulation requires) produces the same result as taking each coordinate step individually. Finally, this process is repeated until convergence.

\subsection{CSMDS Alternatives}
\begin{figure}[htb!]
    \centering
   \begin{subfigure}[h]{0.49\linewidth}
      \includegraphics[width=\linewidth]{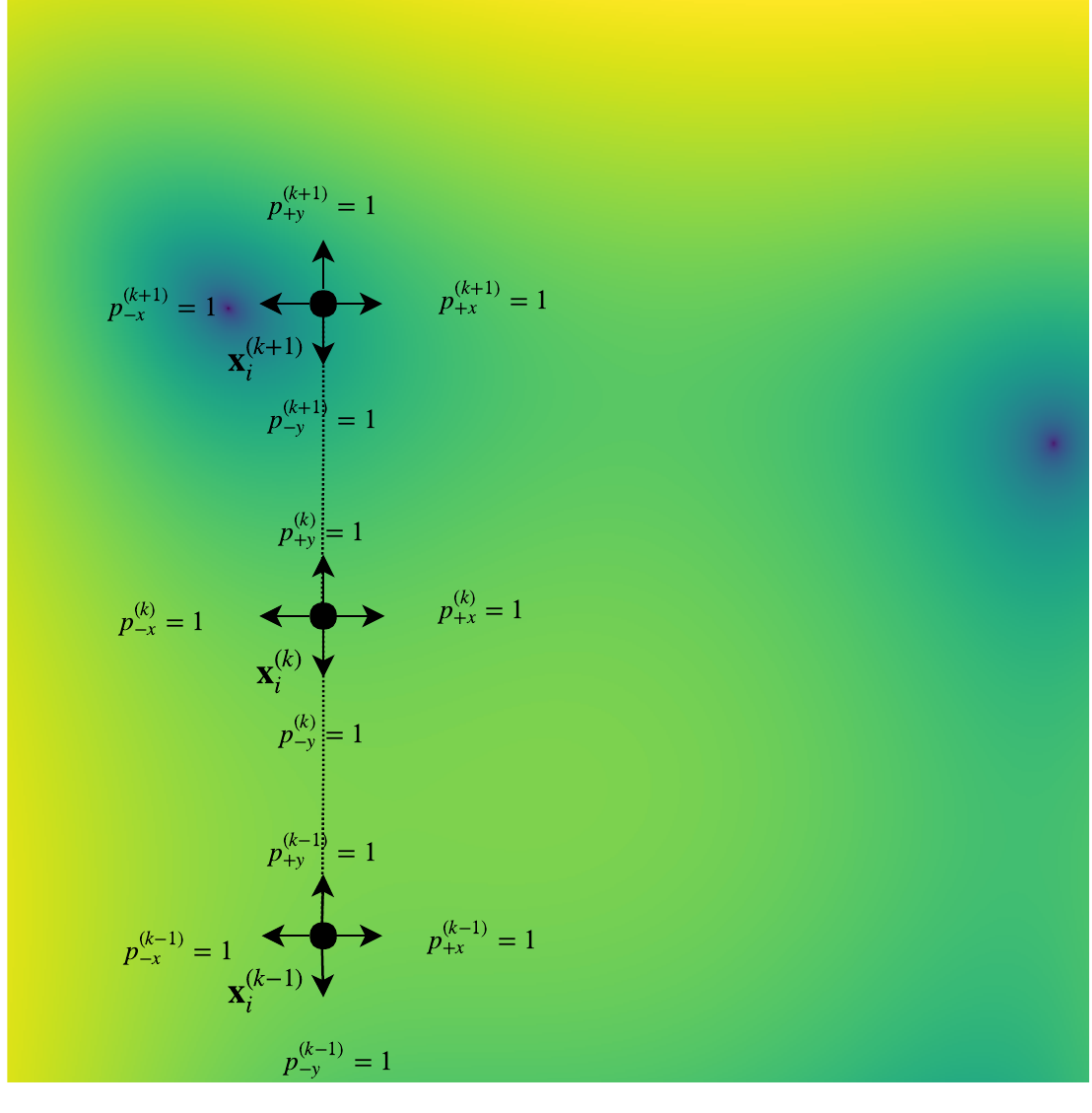}
 	 \caption{Full-Search (FS) CSMDS}
 	  \label{fig:FSCSMDS}
  \end{subfigure}
  \begin{subfigure}[h]{0.485\linewidth}
      \includegraphics[width=\linewidth]{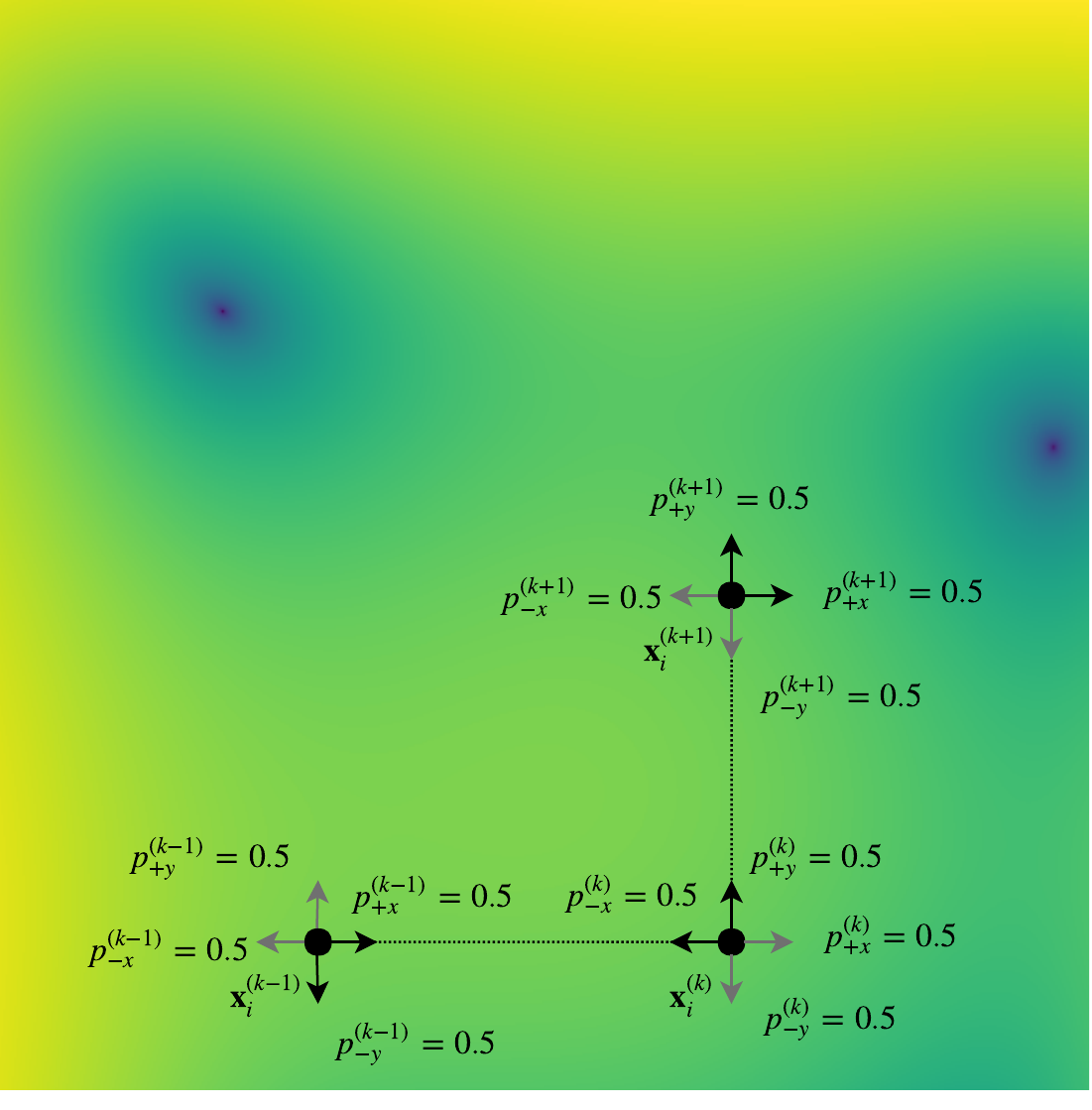}
      \caption{Randomized (RN) CSMDS}
      \label{fig:RNCSMDS}
  \end{subfigure} \\
  \begin{subfigure}[h]{0.7\linewidth}
      \includegraphics[width=\linewidth]{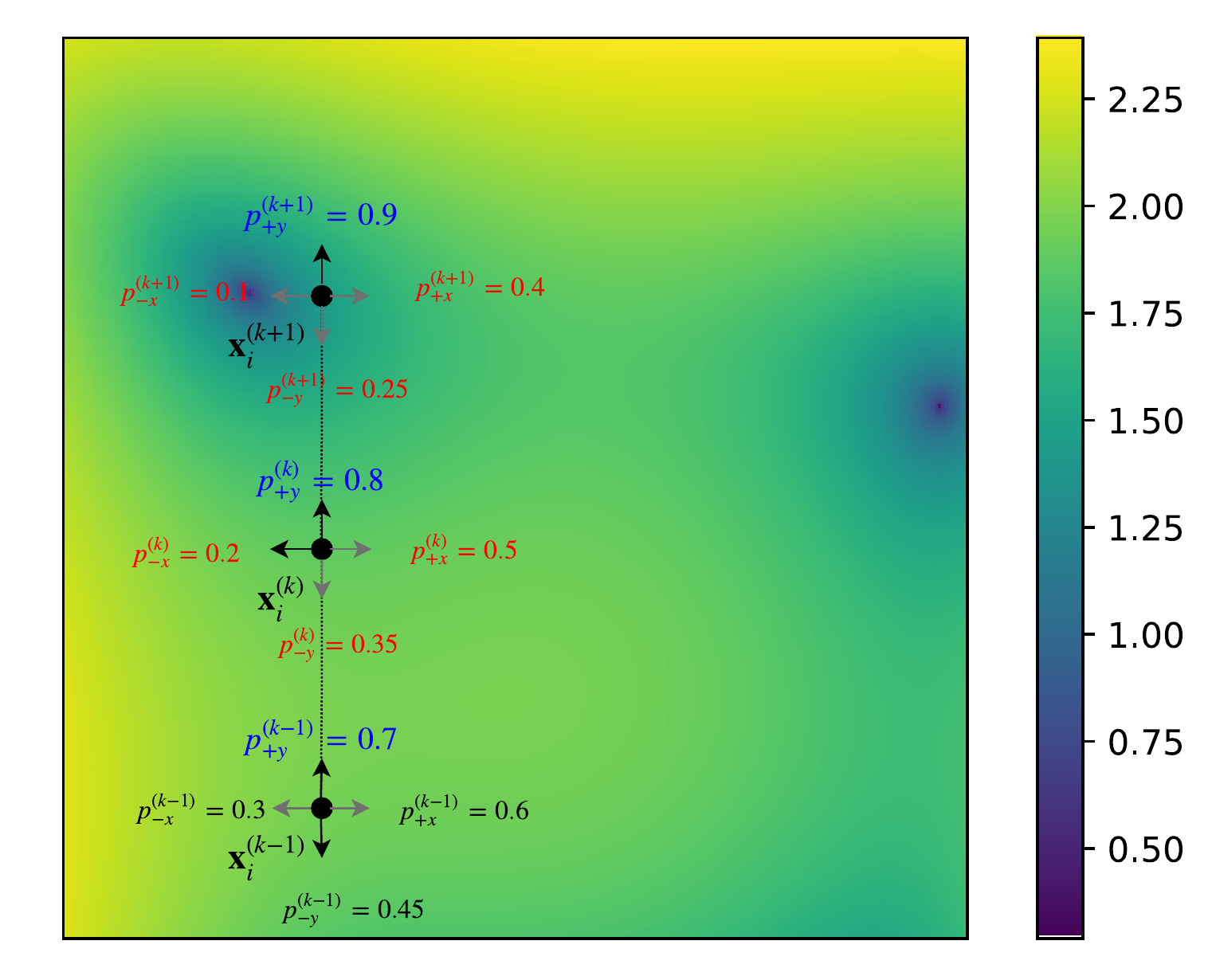}
      \caption{Bootstrapped (BS) CSMDS}
      \label{fig:BSCSMDS}
  \end{subfigure}
    \caption{Example of how CSMDS proceeds for one point $\textbf{x}_i \in \mathbb{R}^2$ in order to minimize the objective function for different alternatives. Black color arrows denote the coordinates that were actually evaluated towards the minimization for this epoch while gray ones denote the coordinates that were omitted (the corresponding Bernoulli variable was unsuccessful). a) Full-Search would always consider each possible coordinate with probability $1$ and thus the objective function would be evaluated for all of them. Of course this would give us the best possible solution for each iteration. b) Randomized alternative would have a constant probability for searching all over the directions. This might result to the optimal coordinate step to be omitted from the evaluation. c) Bootstrapped CSMDS enhances the probability of the optimal coordinate from the previous epoch (Blue colored probabilities) and reduces the probabilities of all the other coordinates (Red colored probabilities). In this way, it becomes more likely to continue to move towards a direction that produces a decrease of the objective function while also reducing the computational cost for searching directions which were not producing the steepest decrease on previous iterations. So we might not need to evaluate the latter coordinates at all in many cases.}
    \label{fig:CSMDSalternatives}
\end{figure}

We will distinguish three different cases of the CSMDS general framework that was described earlier where each case can be configured by giving different configurations of the initial probability matrix $\tilde{\mathbf{P}}^{(0)}$, the threshold for each entry of the latter probability matrix $p_{th}$ and the probability update step $p_a$ which are the calling parameters of Algorithm \ref{a:proposed}. For the cases of Full-Search (FS) and Randomized (RN) CSMDS the probability matrix would not be updated across epochs but they would be specified by the given probability matrix $\tilde{\mathbf{P}}^{(0)} \in [0,1]^{N \times 2L}$. Thus, each positive or negative coordinate would correspond to a Bernoulli random variable with a constant probability. On the contrary, $\tilde{\mathbf{P}}^{(k)}$ would be updated for each epoch by increasing the probability of the coordinates that produce the optimal decrease and by reducing all the other. In Figure \ref{fig:CSMDSalternatives} we provide a visualization of how the algorithm proceeds for the three different alternatives of CSMDS.

\subsubsection{Full-Search CSMDS (FS CSMDS)}
FS CSMDS can be configured as an instance of the unified CSMDS framework (see Algorithm \ref{a:proposed}) with an initial probability matrix $\tilde{\mathbf{P}}^{(0)}$ which has one in all of its entries and with no updates on this matrix which is equivalent to specifying the argument $p_a = 0$. Of course this configuration would always give the best results from all kind of randomized alternatives that can be modeled using CSMDS framework but this comes with a huge computational cost because in every epoch we will consider all possible directions even if some of them have a history of being completely useless towards the minimization of the objective function (see example in Figure \ref{fig:FSCSMDS}). The complexity of FS CSMDS is $\mathcal{O}(N^2L)$. Because for each epoch we search across $2L$ dimensions for $N$ points while we also need $\mathcal{O}(N)$ operations to update the distance matrix as we move all points independently for each epoch.

On the contrary, this variant of CSMDS framework comes with a theoretical convergence property because in each epoch we search all over the possible coordinates. In \cite{paraskevopoulostzinis2018MDS} this has been proven because FS CSMDS is an instance of GPS methods as we have already shown for the general CSMDS framework in Section \ref{MDS:GPSFormulation}. The core idea of the proof is that because FS CSMDS is an instance of GPS methods the we can use Theorem \ref{theorem: Mild Convergence} because the following are true: 

\begin{enumerate}
    \item The stress objective function $g$ (see Equation \ref{eq MSE of the vectorized var}) is continuously differentiable around all points of the search space $\mathbb{R}^{N \cdot L}$ as it is a sum of norms except of the points where we have the minimization $\mathbf{x}_i = \mathbf{x}_j$. Thus, we can relax the initial requirements of the theorem on the set of open balls $\underset{{\mathbf{a} \in L(\mathbf{z}^{(0)})}}{\bigcup} B(\mathbf{a}, \eta)$ where $\eta > 0$ by alternating the set of minimizers $\mathcal{Z}_{\star}$ in order to include the points of $L(\mathbf{z}_0)$ where $\mathbf{x}_i = \mathbf{x}_j$ \cite{torczon1991convergence}.
    \item Pattern matrix $\hat{\mathbf{P}}$ in Equation~\ref{eq Our algorithm's Pk} contains all the possible coordinate step vectors (which are also defined by Equation~\ref{eq GPS trial step}) which are evaluated with probability vector $1$ for each point and  epoch. Thus, if there exists a simple decrease when moving towards any of the directions provided by the columns of $\Psi$ then our algorithm also provides a simple decrease. As a result, Hypothesis~\ref{hyp: mild exploratory moves} holds for the exploratory coordinates.
\end{enumerate}

By combining the latter two intermediate results, Theorem~\ref{theorem: Mild Convergence} holds for FS CSMDS and thus, $\underset{k \rightarrow +\infty}{\lim} \inf \; ||\nabla g(\mathbf{z}^{(k)})|| = 0$ is guaranteed.

\subsubsection{Randomized CSMDS (RN CSMDS)}
In this case, Algorithm \ref{a:proposed} is configured with an constant probability matrix $\tilde{\mathbf{P}}^{(0)}$ which has the same value $p_{init}$ for all of its entries but it should be less than one in order to avoid evaluating every time all possible coordinates. Moreover, because there is no update on the probability matrix $p_a = 0$. Intuitively, RN CSMDS takes advantage of some directions that might not be needed to taken into consideration when trying to infer the optimal movement mainly because another direction could also produce a reasonably good decrease. In \cite{paraskevopoulostzinis2018MDS} authors have noted this method as an optimization over FS CSMDS because they have empirically seen that this produces faster convergence. However, they have not explicitly taken into consideration its proper formalization under a general framework. Now the complexity of RN CSMDS is $\mathcal{O}(N^2R)$ where $R = \lceil{L \cdot p_{init}}\rceil$ is the expected number of search coordinates that the algorithm would evaluate with $R<L$. The problem is that by just random sampling over the possible coordinates does not always perform favorably as sometimes we might need to keep moving towards a specific direction for multiple epochs but there is no structure of history tracking. This might result to unwanted behavior where all coordinates are equally probable to be evaluated and thus, a wrong step is the same probable to happen as the best one (see Figure \ref{fig:RNCSMDS}).

\subsubsection{Proposed Bootstrapped CSMDS (BS CSMDS)}
In the proposed Bootstrapped (BS) CSMDS we try to incorporate an importance sampling scheme in order to favor the coordinates that have previously produced optimal decrease of the loss function over all the other coordinates. This method would presumably help in certain cases where we only need to move along specific directions for certain epochs (see example in Figure \ref{fig:BSCSMDS}). BS CSMDS inherently adds an equivalent inertia term for the coordinate search algorithm as more contemporary gradient-based optimizers also do, e.g. Nadam optimizer \cite{nadamOptimizer}. For the configuration of BS CSMDS we also need probability matrix $\tilde{\mathbf{P}}^{(0)}$ which would be updated over time, a threshold probability $p_{th} \in [0, 1]$ and a non-zero update probability step $p_a \in (0,1]$. 

The number of the coordinates that would be evaluated for each point would decrease over time. Let us assume without loss of generality that all values of the initial probability matrix $\tilde{\mathbf{P}}^{(0)}$ are set to $p_{init} \in (0,1]$ and a given threshold probability that we cannot reduce any probability entry below that $p_{th} < p_{init}$. Omitting the zero-step epochs for certain points that do not produce any update on the probability matrix, in every epoch we only increase the probability of the optimal direction by $p_a$ and for all the other $2L-1$ we reduce it by the same amount. So for each successful iteration of a certain point the number of coordinates would be reduced by $\frac{(2L-2)\cdot p_a}{2L}$. So as epochs goes to infinity, the probability of evaluating any coordinate would reduce to the threshold value. Thus, the number of the coordinates to be evaluated would be equal to $R_{BS} = \lceil{L \cdot p_{th}}\rceil$. This yields the problem that if the threshold is set extremely low this saturation happens very fast then we might end up with not being able to search over any coordinate that would actually provide a decrease of the objective function. So one might consider that it would be wise to select a threshold $p_{th}$ which would enable the algorithm to search over coordinates even when it trespasses over the saturation phase. Notably, BS CSMDS is able to self-regulate even in saturation-phase if there is a decrease over a certain coordinate. In essence, BSMDS will evaluate this coordinate with probability $p_{th}$ (in the worst case) and it will iteratively increase this probability by $p_a$ for each successful iteration.     

\subsection{Implementation Details}
We provide a highly efficient Python code for CSMDS which runs C and Cython routines which are computationally intense. Moreover, our implementation is highly parallelizable as each evaluation across a possible coordinate can run on a different thread and it is completely independent from all the other coordinates.

\section{Experiments}
We assess the performance of the proposed BS CSMDS using both synthetic and real image data. We compare with the state-of-the-art SMACOF MDS but also for the other two alternatives of CSMDS framework as described in the previous section. For the configuration of the starting radius parameter $r^{(0)}$ we select the value of $5$ empirically as it is self-regulating in all of the cases because if it does not produce a relative error decrease better than $\epsilon = 10^{-4}$ between two consecutive epochs then radius will be reduced by $50\%$. The parameter for checking the convergence threshold or equivalently the threshold of how small the searching radius would be is set to $\delta = 10^{-3}$. 

\subsection{Experiments with Synthetic Data}
\label{sec:SynthDataExperiments}
\begin{figure}[htb!]
    \centering
   \begin{subfigure}[h]{0.59\linewidth}
      \includegraphics[width=\linewidth]{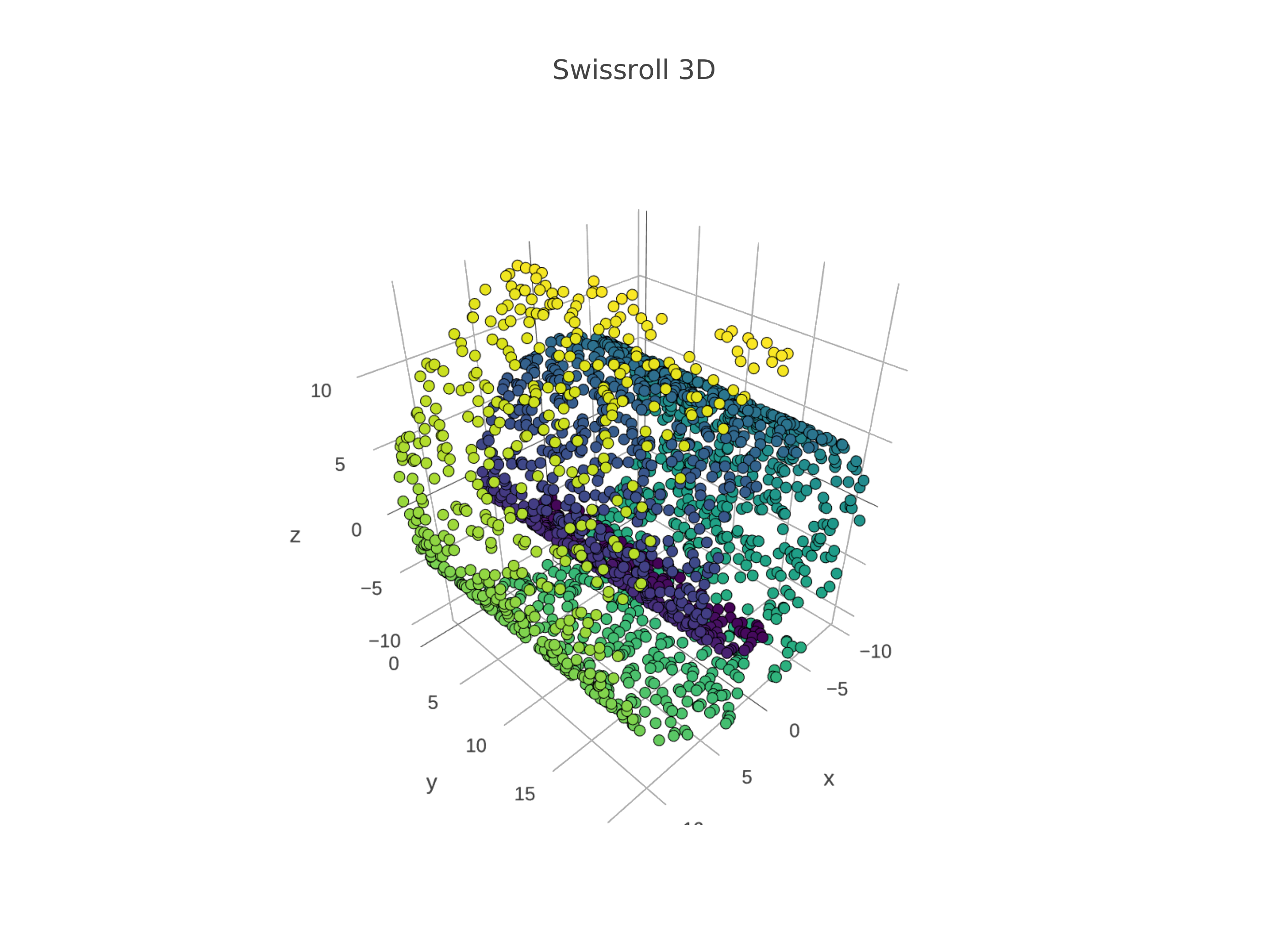}
 	 \caption{Original Swissroll 3D}
 	  \label{fig:swissroll3d}
  \end{subfigure} \\
  \begin{subfigure}[h]{0.49\linewidth}
      \includegraphics[width=\linewidth]{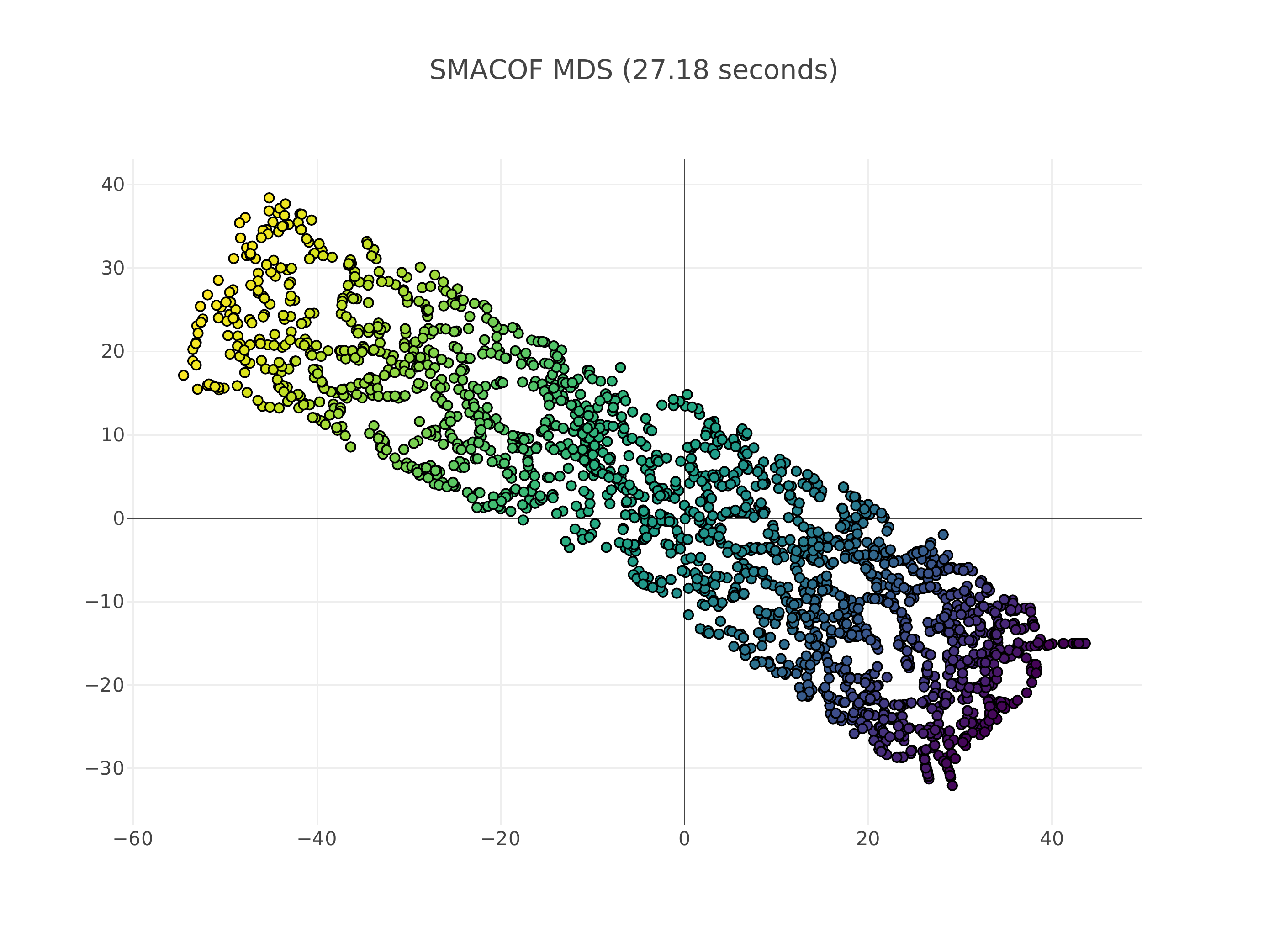}
      \caption{SMACOF MDS}
      \label{fig:swissSMACOF}
  \end{subfigure} 
  \begin{subfigure}[h]{0.49\linewidth}
      \includegraphics[width=\linewidth]{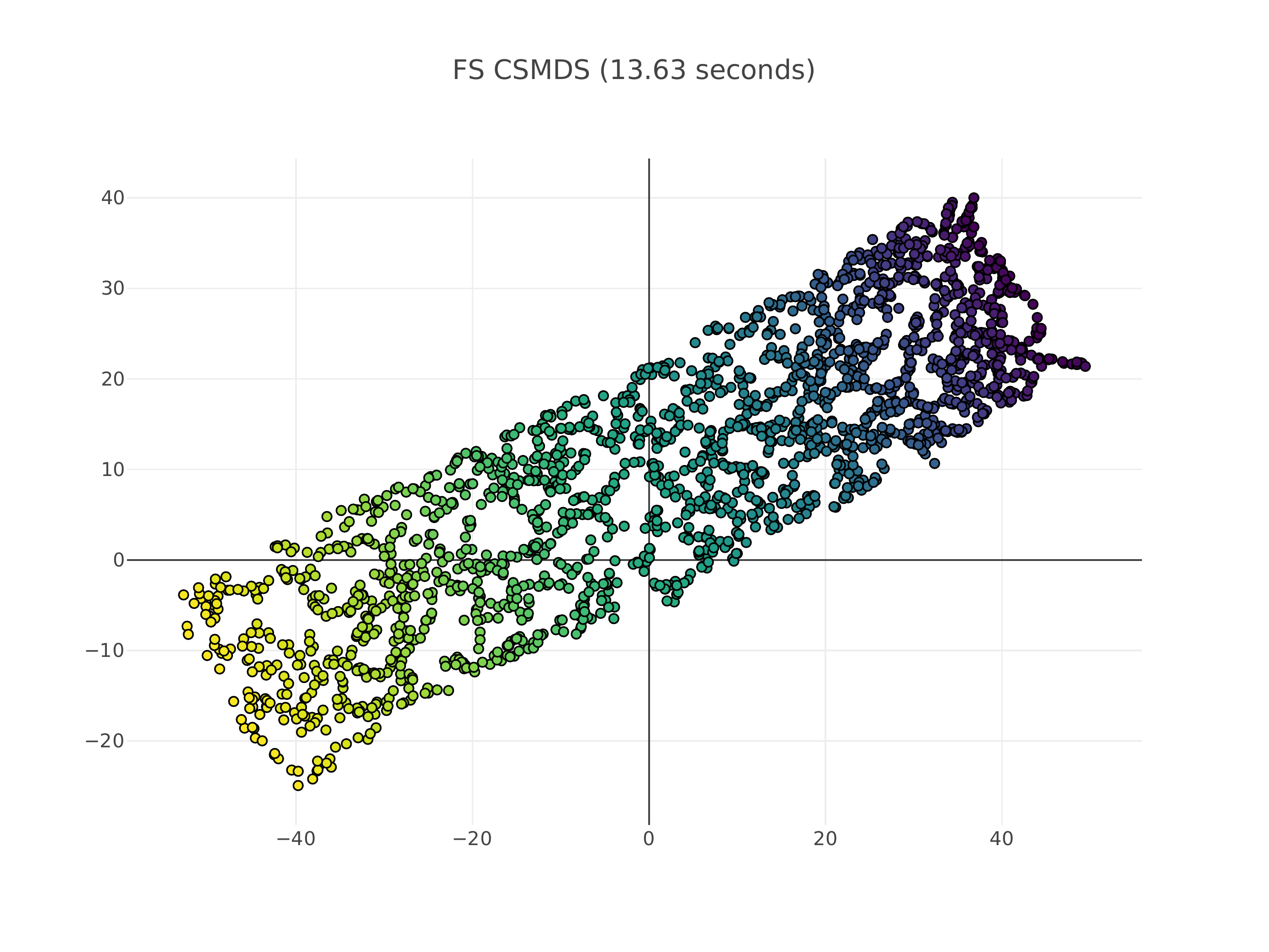}
      \caption{Full-Search (FS) CSMDS}
      \label{fig:swissFSCSMDS}
  \end{subfigure} \\
  \begin{subfigure}[h]{0.49\linewidth}
      \includegraphics[width=\linewidth]{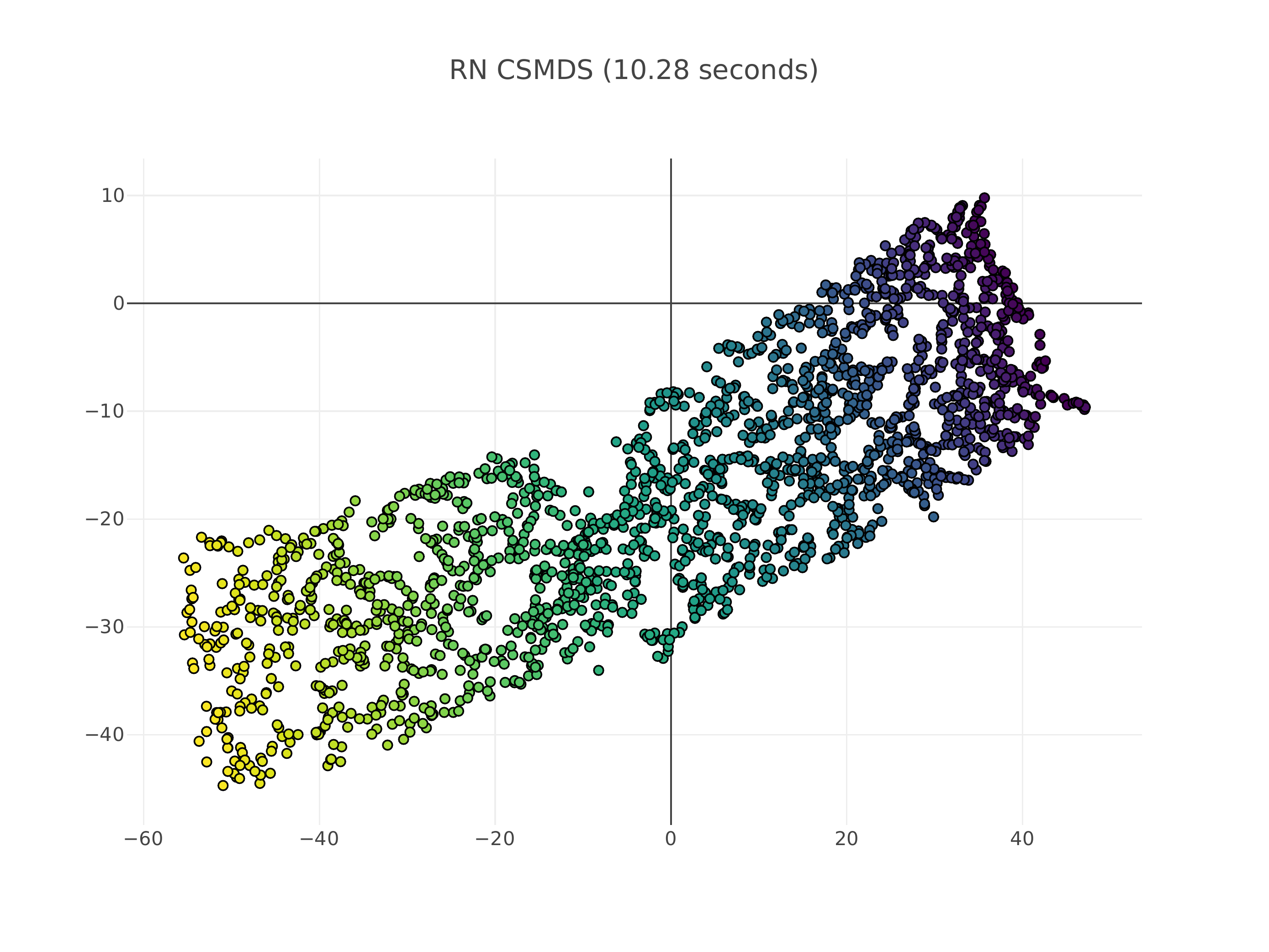}
      \caption{Randomized (RN) CSMDS}
      \label{fig:swissRNCSMDS}
  \end{subfigure} 
  \begin{subfigure}[h]{0.49\linewidth}
      \includegraphics[width=\linewidth]{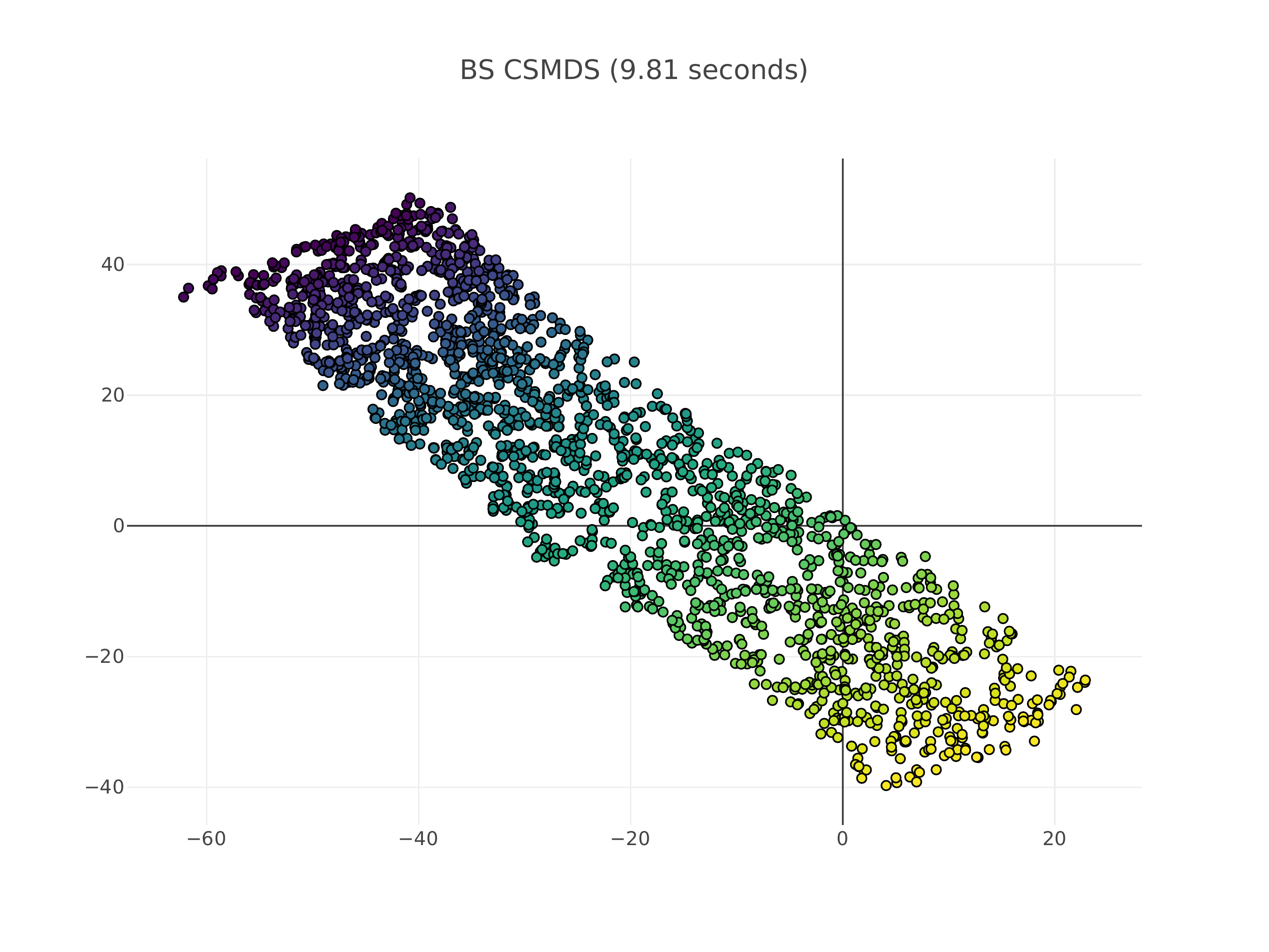}
      \caption{Bootstrapped (BS) CSMDS}
      \label{fig:swissBSCSMDS}
  \end{subfigure}
    \caption{Comparison of Linear embeddings obtained by SMACOF and CSMDS alternatives for the reconstruction of the geometry of swissroll data embedded in $\mathbb{R}^3$ by considering as target matrix the matrix of the geodesic distances obtained by running Bellman-Ford algorithm on the graph of the three-dimensional points. Notice that BS CSMDS is able to preserve the geometry of the manifold on the low-dimensional space with a much lesser computational time compared to all the other variants of CSMDS framework but also compared to the MDS state-of-the-art SMACOF algorithm.}
    \label{fig:Synthetic}
\end{figure}

In this experimental setup we perform non-linear dimensionality reduction using MDS. This can be done in the same way as ISOMAP \cite{tenenbaum_global_2000} first extracts the geodesic distance matrix by considering high-dimensional points as graph nodes and then connecting each node with a number of nearest neighbors and the value of the edge is the corresponding Euclidean distance. The assumption here is that data lie on a manifold which is locally Euclidean and thus geodesic distances would preserve the true structure of the data when performing MDS. In Figure \ref{fig:Synthetic} we compare BS CSMDS with the FS and RN variants as well as SMACOF MDS which is considered the state-of-the-art algorithm for MDS when generating $2000$ $3D$ data points which are drawn from a Swissroll distribution. All algorithms run for $200$ epochs and converged to same level of Stress errors, while SMACOF yielded a higher error compared to all other CSMDS variants. RN CSMDS and BS CSMDS are configured with $p_{init}=0.7$, while the latter uses the extra arguments: $p_a=0.05$ and $p_{th}=0.2$. Experimentally, we noticed that when using a much higher probability update $p_a$ we very quickly force the algorithm to enter the saturating phase which produces weird error rates. 

It is very important to notice that all algorithms aptly capture the intrinsic geometry of the data (which is actually $2$-dimensional and thus, it can be visualized perfectly when unfolding Swissroll using only $2$ dimensions) when the target matrix includes the geodesic distances between each pair of local neighborhood points. Moreover, we can clearly see that from the structure of the low-dimensional representations all MDS algorithms under comparison are converging to similar solutions besides the way we choose to optimize the Stress function by Coordinate Search that CSMDS variants use or either minimization by majorization that SMACOF uses. Another important conclusion that we draw is that when using a small number of dimensions in order to describe our data, CSMDS converge way faster than SMACOF to an acceptable error. This certainly would not be the case when one uses much more dimensions because the computational time of all CSMDS variants scale linearly with the number of dimensions which are used (CSMDS variants would have different constants depending on the expected number of dimensions that we evaluate for each epoch). On the contrary, complexity of SMACOF does not depend on the number of dimensions but generally need a number of runs larger than one in order to find a good solution \cite{de2011multidimensional}, in this case we only use one run that produces an acceptable solution in order to be fair in time-comparison. Having all that said, we can clearly see that CSMDS variants are extremely wants to perform MDS with a small number of dimensions. Finally, the proposed CSMDS is able to outperform all other variants even in this case with $2$-dimensions which is somewhat unintuitive because one would expect to see an unwanted behavior when we only have $4$ coordinates to move for each point (all of them might be needed).

\subsection{Experiments with MNIST}
In this experimental setup we perform linear dimensionality reduction using as target dissimilarity matrices the Euclidean ones from the high-dimensional pixel space $\mathbb{R}^{784}$ of MNIST dataset \cite{lecun1998mnist}. 

\subsubsection{Visualizing Linear Embeddings}
\label{sec:visualizingmnist}
\begin{figure}[htb!]
    \centering
  \begin{subfigure}[h]{0.49\linewidth}
      \includegraphics[width=\linewidth]{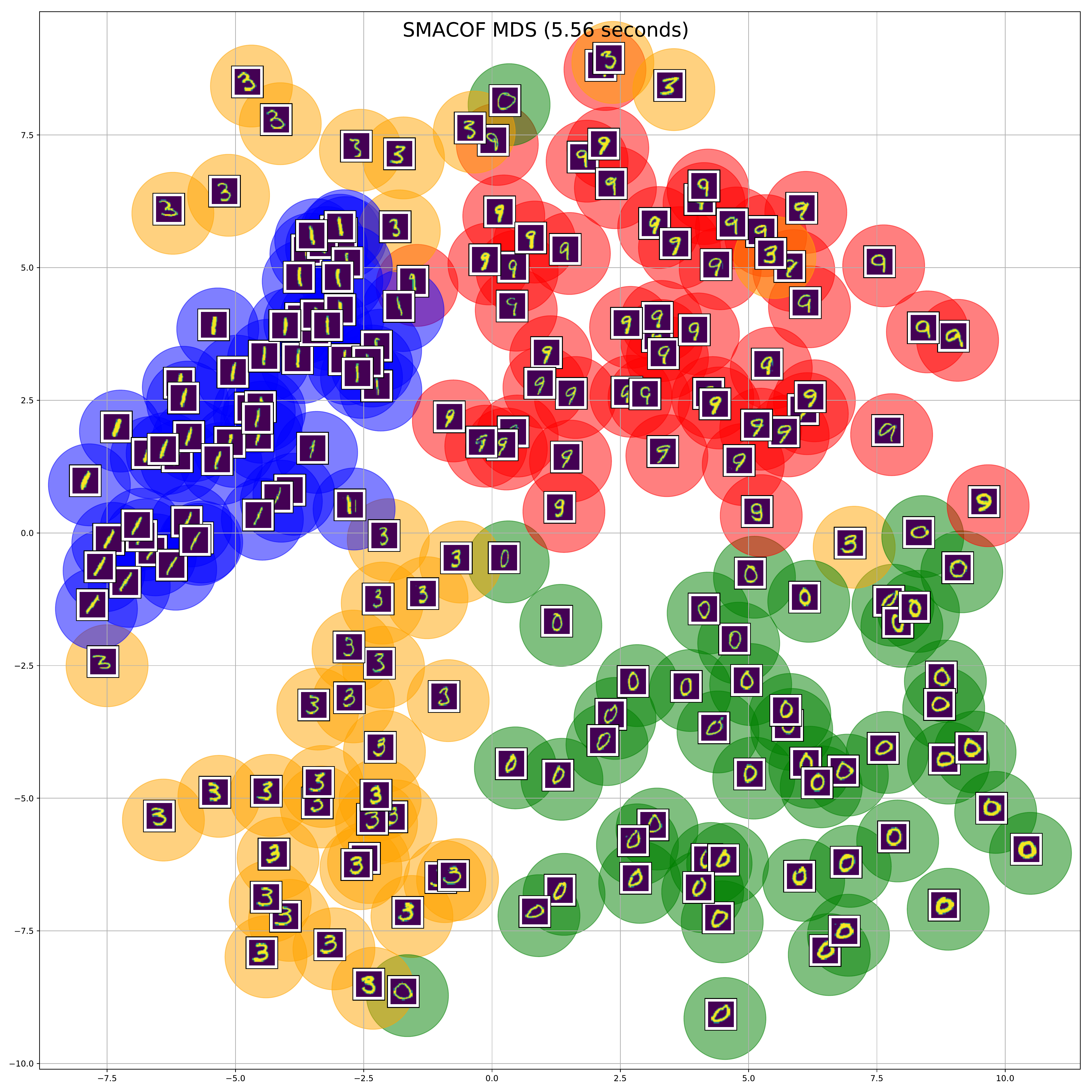}
      \caption{SMACOF MDS}
      \label{fig:mnistSMACOF}
  \end{subfigure} 
  \begin{subfigure}[h]{0.49\linewidth}
      \includegraphics[width=\linewidth]{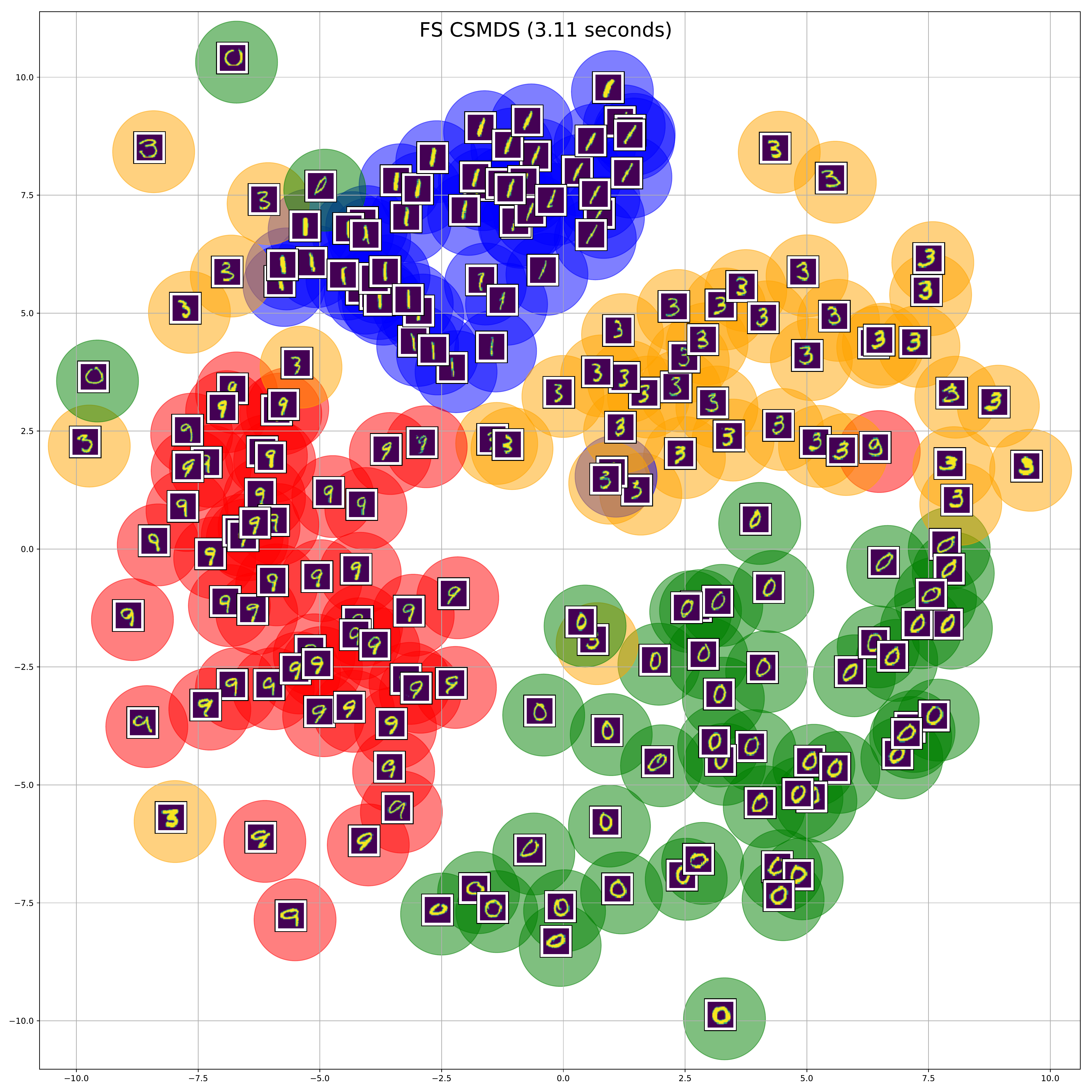}
      \caption{Full-Search (FS) CSMDS}
      \label{fig:mnistFSCSMDS}
  \end{subfigure} \\
  \begin{subfigure}[h]{0.49\linewidth}
      \includegraphics[width=\linewidth]{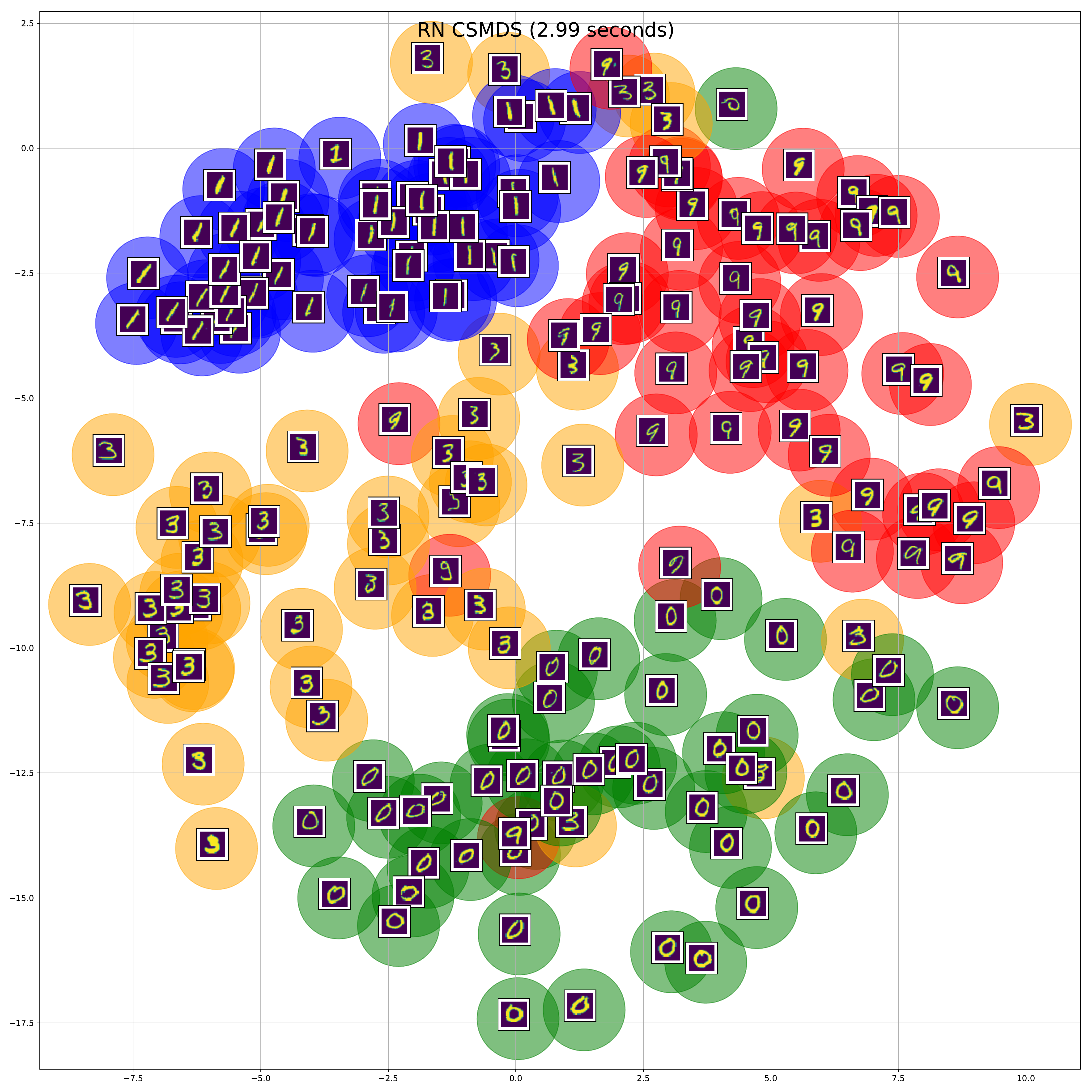}
      \caption{Randomized (RN) CSMDS}
      \label{fig:mnistRNCSMDS}
  \end{subfigure} 
  \begin{subfigure}[h]{0.49\linewidth}
      \includegraphics[width=\linewidth]{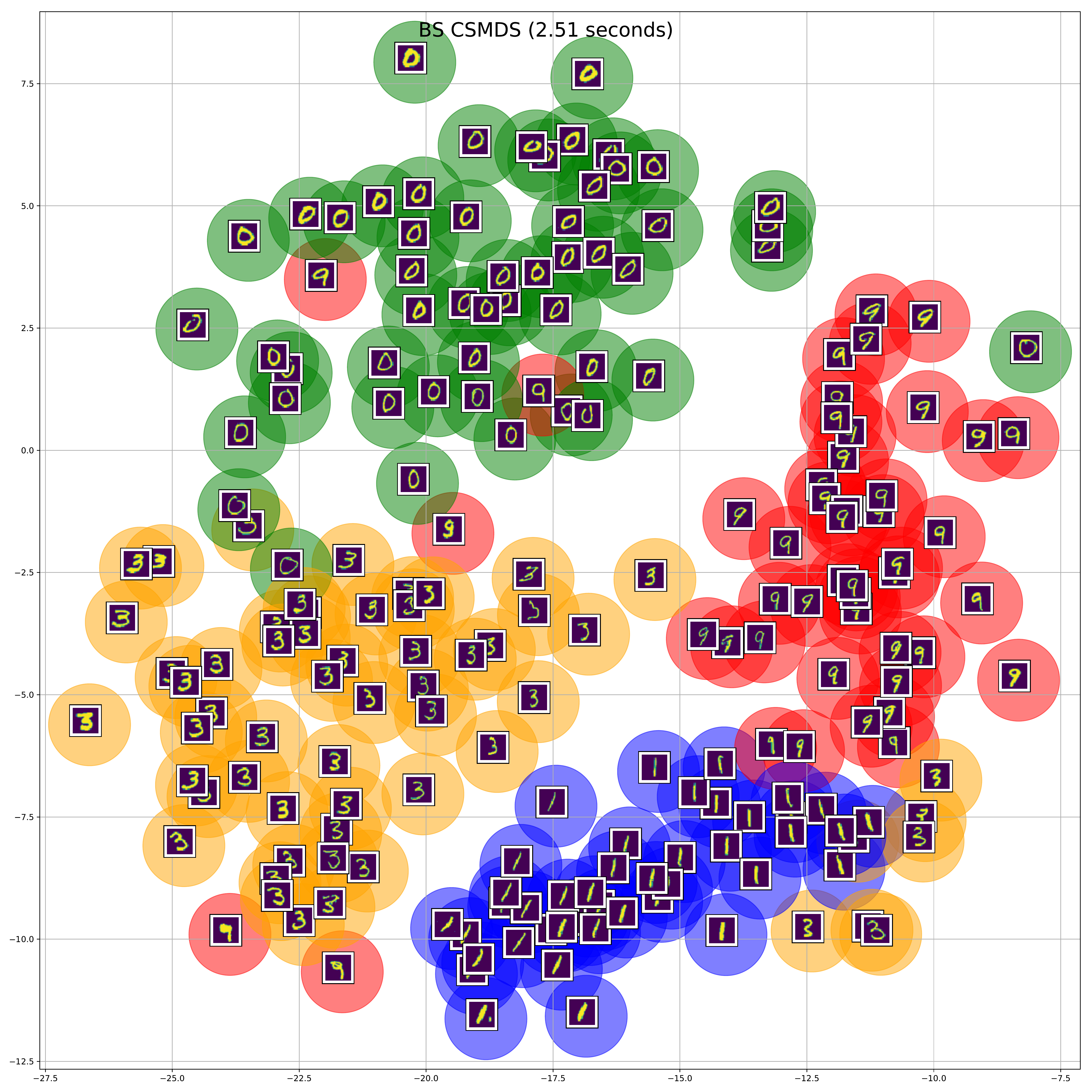}
      \caption{Bootstrapped (BS) CSMDS}
      \label{fig:mnistBSCSMDS}
  \end{subfigure}
    \caption{Comparison of linear-dimensional manifold reconstruction obtained by SMACOF and CSMDS alternatives from MNIST digits considering as target dissimilarity matrix the Euclidean space on the pixel space $\mathbb{R}^{784}$. Equivalent solutions are obtained from all MDS optimization algorithms. Proposed BS CSMDS yields a centered clustered reconstruction for each digit in less than one half of the time that state-of-the-art SMACOF MDS needs. Moreover, it outperforms all other CSMDS alternatives even in this case where the total number of available coordinates is very small.}
    \label{fig:visualizingMNIST}
\end{figure}

We use only $4$ out of the $10$ available handwritten digits, namely: $\{0,1,3,9\}$ and we select randomly $1000$ images from all the selected classes of digits. Moreover, we compute the Euclidean distance matrix between all the images on the pixel space. After that, we perform MDS using SMACOF and all variants of CSMDS using only $2$-dimensions. All algorithms run for $200$ epochs and converged to same level of Stress errors. RN CSMDS and BS CSMDS are configured with $p_{init}=0.7$, while the latter uses the extra arguments: $p_a=0.05$ and $p_{th}=0.2$. The resulting embeddings after some random sampling on the constructed low-dimensional space (for visualization purposes) are displayed in Figure \ref{fig:visualizingMNIST}. Different colors correspond to the classes of different handwritten-digits $\{0,1,3,9\}$ which are mapped to $\{$\textit{green}, \textit{blue}, \textit{orange}, \textit{red}$\}$, correspondingly.  

From the visualization of the embeddings we can clearly see that all MDS algorithms are able to preserve the geometry of the hand-written digits in a way that all digits of the same class would lie on compact $2D$ Euclidean subspaces. Although, RN CSMDS provides a good embedding, it performs poorly for the class of digit $3$ which has the most variance among all algorithms. On the contrary, again we see that we get a similar result with the experiments on synthetic data (see Section \ref{sec:SynthDataExperiments}) where BS CSMDS provides a very good solution by outperforming SMACOF and the other CSMDS variants in terms of computational efficiency. Noticeably, BS CSMDS constructs a very dense and center-based cluster representation even for the class of digit $3$ which makes this representation amenable for center-based clustering and classification.     

\subsubsection{Image Classification on the Low-Dimensional Space}
We steer our focus on how the embeddings of the different MDS algorithms behave under a true classification experimental setup. For this experiment all classes of $10$ handwritten digits have been used and $3000$ images have been drawn randomly from all these classes which have a very similar prior distribution on the initial dataset. We first perform linear-dimensionality reduction using MDS with target dissimilarity matrix the euclidean distances on the pixel space in order to obtain $10$-dimensional embeddings. We randomly spit the embeddings in two sets where the labels of $90\%$ of the images ($2700$) would be considered known and used in order to infer the class-labels of the remaining $300$. On the produced embeddings a K-Nearest Neighbor (KNN) classification is performed using different parameters of nearest neighbors which are taken into consideration for the inference of a sample which comes from the test set. The evaluation metric which is considered is the total classification accuracy which is the percentage of true predictions across all classes divided by the number of total predictions. In order to get a better insight of the accuracy which is obtained by these embedding representations we also perform KNN on the orignal space without performing any dimensionality reduction (and possibly loose some information which is crucial for classification). RN CSMDS is configured with $p_{init}^{RN}=0.7$, while the BS CSMDS is configured with: $p_{init}^{BS}=0.4$, $p_a=0.05$ and $p_{th}=0.2$. The results for both the computational time which was needed in order to obtain the embeddings and the final accuracy from KNN classification is shown in Table \ref{t:mnist}. 
\begin{table}[htb!]
\centering
\begin{tabular}{lccccccc}
\toprule
Method &  Dims &    Time (secs) &    $K=1$ &    $K=3$ &    $K=5$ &   $K=7$ &    $K=9$ \\
\midrule
Initial            &   784 &    - &  93.67 &  94.00 &  94.33 &  94.00 &  94.00 \\
\midrule
SMACOF MDS         &    10 &  102.64 &  84.00 &  85.00 &  86.00 &  88.00 &  86.33 \\
FS CSMDS  &    10 &  185.83 &  \textbf{89.33} &  \textbf{90.67} &  91.33 &  90.67 &  \textbf{92.33} \\
RN CSMDS   &    10 &  136.31 &  89.00 &  89.67 &  91.00 &  90.67 &  91.33 \\
BS CSMDS &    10 &   \textbf{85.31} &  86.67 &  88.67 &  \textbf{91.67} &  \textbf{91.00} &  91.33 \\
\bottomrule
\end{tabular}
\vspace{0.2cm}
\caption{Comparison of MDS linear dimensionality reduction techniques for the MNIST dataset with embedding dimension equal to 10 for the classification using KNN algorithm. Parameter $K$ is the number of nearest neighbors which are taken into consideartion from the training set in order to infer the class-label for an embedding of the test set. ``Time" column specifies the amount of time that each method took to compute all embedding vectors. ``Initial" corresponds to performing KNN on the original high-dimensional pixel space $\mathbb{R}^{784}$.}
\label{t:mnist}
\end{table}

Even now that we have increased significantly the number of dimensions of the embedding space again we see that BS CSMDS outperforms all other alternatives and SMACOF in terms of computational efficiency but without having a significant performance drop in terms of classification accuracy on the embedded space. Compared to FS CSMDS which obtains the best results under most configurations but also is the most computationally intensive task, BS CSMDS manages to obtain similar classification accuracy in all of the cases while sometimes even surpassing FS alternative. Moreover, RN CSMDS always has a higher probability $p_{init}^{RN}=0.7$ of searching a possible coordinate to optimize compared to proposed BS CSMDS ($p_{init}^{BS}=0.4$) but does not yield any better results in terms of classification accuracy and of course it is much more computationally intensive. This indicates that only a random sampling across the dimensions we want to optimize is not sufficient in order to significantly reduce the computational time of CSMDS algorithms and also obtain a ``good" embedding representation which is amenable to classification and clustering. Presumably, only a small number of all the available coordinates to be searched is needed even in the case of $20$ possible directions.    

\subsubsection{Head-to-Head Convergence Comparison for CSMDS Alternatives}
\begin{figure}[htb!]
    \centering
    \includegraphics[width=\linewidth]{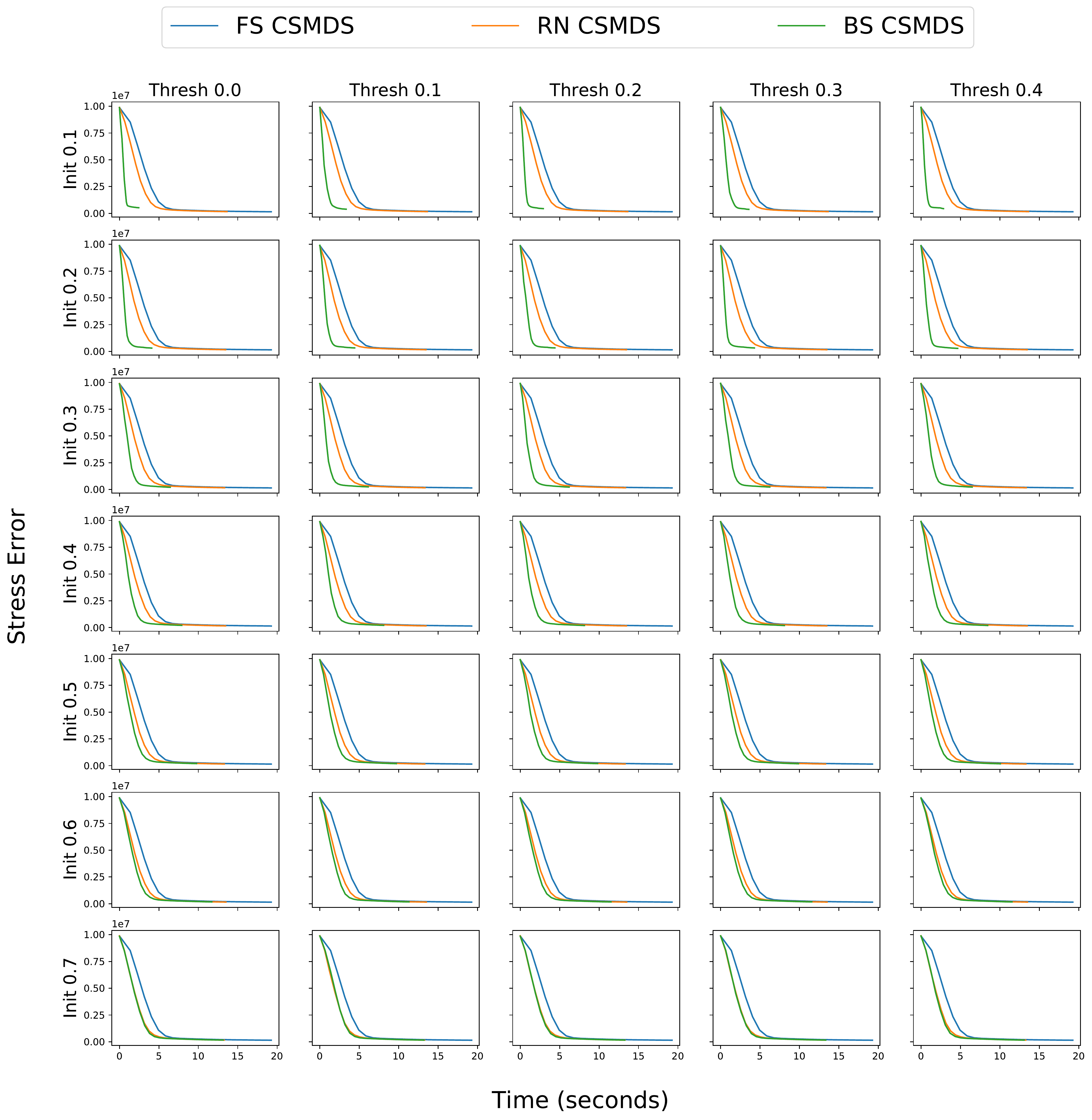}
    \caption{Convergence evaluation for all variants of CSMDS unified framework when performing linear dimensionality reduction on MNIST images from the original space $\mathbb{R}^{784}$ to the embedding space $\mathbb{R}^{100}$. For each row we alter accordingly the initial probability matrix parameter $p_{init}$ and across columns we configure accordingly the probability threshold parameter $p_{th}$. }
    \label{fig:performance}
\end{figure}

In this section we aim to extensively evaluate all variants of CSMDS when performing dimensionality reduction using a much higher number of embedding dimensions. We also focus on understanding how the proposed BS CSMDS behaves under different configurations of the initial probability $p_{init}$ and the probability threshold $p_{th}$ which defines the behavior of the bootstrapped version on the saturating phase. In order to do so, we randomly select $1000$ MNIST images from all the available digit-classes and we set the target embedding dimension to be $100$. We keep the same configuration for the probability update step $p_a=0.05$. We measure how the Stress error function (see Equation \ref{eq Our algorithm's minimization}) decreases across time for FS CSMDS, RN CSMDS and proposed BS CSMDS under the aforementioned setup. In Figure \ref{fig:performance}, we plot the value of Stress error function across time for all CSMDS variants under different configurations of $p_{init}$ (shared for the configuration of RN and BS alternatives) and $p_{th}$ (behavior changes only for BS CSMDS). Also, FS CSMDS is not configured by neither of the parameters $p_{init}$ and $p_{th}$. Consequently, FS CSMDS is the same for all plots and RN CSMDS changes only across rows. However, we plot every time all alternatives in order to provide a much easier inspection for the comparison of the performance.     

It is evident that the proposed variant of BS CSMDS yields the best convergence results across all configurations and surpasses RN and FS but with a different margin. That means that BS CSMDS is quite robust and can self-regulate under different configurations. RN CSMDS also converges to approximately the same level as FS CSMDS but faster in all of the cases which is in accordance with the empirical conclusion authors state in \cite{paraskevopoulostzinis2018MDS}. It is important to notice that the latter holds irregardless of the initial probability that is set. This can be explained because in RN alternative the sampling of the searching directions is completely random and thus when we decrease $p_{init}$ the time that we spend for the evaluation of the objective function on each epoch but more epochs are needed in order to converge. On the contrary, BS CSMDS enhances the probability of evaluating directions that yield steepest descent of the objective function and thus it manages to converge much faster than all the other alternatives. This fact becomes much more apparent when configuring BS CSMDS with lower initial probability $p_{init} = 0.1$ or $p_{init} = 0.2$. In these cases BS CSMDS manages to find these optimal directions, enhances the probability of selecting them in later evaluations and thus very quickly manages to reduce the error to the same level as RN and FS alternatives around $5-7$ times faster. As we increase the initial probability $p_{init}$ BS CSMDS becomes very similar to the behavior of RN alternative as the initial probability dominates and in the first $20$ epochs that we run, all coordinates have an equal probability of being evaluated. Thus, BS CSMDS cannot properly enhance the probability of the steepest direction and it has much more random behavior because all directions have approximately the same probability to be evaluated. Finally, we have to identify that CSMDS alternatives irregardless of how quick the converge to a good solution when we want to do smaller movements around a local minima then the computational time which is needed is extremely expensive and consequently, gradient-based solvers are would be much more efficient. However, with BS CSMDS we are able to converge extremely fast (compared to the other CSMDS alternatives) around a local minimizer that could potentially serve as an initialization for a much faster gradient-based optimizer.

\section{Conclusions}
In this work, we have proposed a unified framework for gradient-free Multidimensional Scaling (MDS) based on Coordinate Search (CS) over the embedding space, namely, CSMDS. We have formally described CSMDS framework as an instance of General Pattern Search (GPS) methods that can be used in order to provide theoretical convergence properties of the deterministic Full-Search CSMDS (FS CSMDS) alternative. We propose a Bootstrapped alternative of CSMDS (BS CSMDS) which enhances the probability of the direction that yields the best decrease of the objective function while also reducing the corresponding probability of all the other coordinates. In this way, BS CSMDS manages to quickly identify the directions that are needed towards minimization and avoids evaluating coordinates that might not produce a simple decrease resulting on significant speedup over other deterministic and randomized CSMDS alternatives under experiments on both synthetic and real data. Because BS CSMDS converges very fast to a local minima, we consider a potential future direction to utilize it as an initialization step for gradient-based solvers.   

\printbibliography

\end{document}